\documentclass[final,1p,times,twocolumn]{elsarticle}

\usepackage{amssymb}

\usepackage{lineno}

\usepackage{comment}
\usepackage{subcaption}
\usepackage{booktabs} 
\usepackage{multirow} 
\usepackage{pifont}
\newcommand{\cmark}{\ding{51}}%
\let\oldemptyset\emptyset
\usepackage{url}

\journal{arXiv}

\begin{document}
	
\begin{frontmatter}

\title{dacl1k: Real-World Bridge Damage Dataset Putting Open-Source Data to the Test}

\author[label1]{Johannes Flotzinger}
\author[label2]{Philipp J. Rösch}
\author[label2]{Norbert Oswald}
\author[label1]{Thomas Braml}
		
\affiliation[label1]{organization={Institute for Structural Engineering, University of the Bundeswehr Munich},
			addressline={Werner-Heisenberg-Weg 39}, 
			city={Neubiberg},
			postcode={85577}, 
			state={Bavaria},
			country={Germany}}
\affiliation[label2]{organization={Institute for Distributed Intelligent Systems, University of the Bundeswehr Munich},
			addressline={Werner-Heisenberg-Weg 39}, 
			city={Neubiberg},
			postcode={85577}, 
			state={Bavaria},
			country={Germany}}

\begin{abstract}
Recognising reinforced concrete defects (RCDs) is a crucial element for determining the structural integrity, traffic safety and durability of bridges. However, most of the existing datasets in the RCD domain are derived from a small number of bridges acquired in specific camera poses, lighting conditions and with fixed hardware. These limitations question the usability of models trained on such open-source data in real-world scenarios. 

We address this problem by testing such models on our ``dacl1k'' dataset, a highly diverse RCD dataset for multi-label classification based on building inspections including 1,474 images. Thereby, we trained the models on different combinations of open-source data (meta datasets) which were subsequently evaluated both extrinsically and intrinsically. 
During extrinsic evaluation, we report metrics on dacl1k and the meta datasets. The performance analysis on dacl1k shows practical usability of the meta data, where the best model shows an Exact Match Ratio of 32\%. Additionally, we conduct an intrinsic evaluation by clustering the bottleneck features of the best model derived from the extrinsic evaluation in order to find out, if the model has learned distinguishing datasets or the classes (RCDs) which is the aspired goal.
The dacl1k dataset and our trained models will be made publicly available, enabling researchers and practitioners to put their models to the real-world test. 
\end{abstract}

		
\end{frontmatter}
	
	
\section{Introduction}
\label{sec:intro}

Against the backdrop of an ageing structure stock as well as the steady increase in heavy traffic, regular and high-quality bridge inspections are indispensable. Simultaneously, affected countries hold to inspection processes that are out-of-date while being confronted with staff shortages.
The final goal of building inspections is the building assessment included in the inspection report. Within this document the damage-informations and -valuations as well as consequential actions (e.g. restoration works, traffic load limitations or bridge closures) are recorded. The recommended actions are determined by the damage-valuation which is based on the damage-information, in addition to the inspector's expertise. Thus, the damage-information is the decisive element. Thereby, each visually and acoustically (Hollowareas) recognisable defect is classified, measured and localised.   
However, in accomplishing this task, the use of computer vision approaches for acquiring the defect-information, within the framework of digitised inspections (DIs), offers great potential for improvement in terms of cost-effectiveness and quality control. 

Major contributions in the field of damage recognition on built structures were made with the advent of datasets for the task of binary \cite{Dorafshan2018Dec,Huthwohl2018Aug,Xu2019,Li2019}, multi-class \cite{Huthwohl2019, Bianchi2021}, multi-label classification \cite{Mundt_2019_CVPR} as well as object detection \cite{Mundt_2019_CVPR} and semantic segmentation \cite{benz2022image, UAV75, CrackSeg9k}. 
Kulkarni~et al.~\cite{CrackSeg9k} combined multiple image datasets of cracked and uncracked surfaces which is, to the best of our knowledge, the only work intersecting with the RCD domain, that is making use of a dataset compilation.

In general, the research area of reinforced concrete damage (RCD) recognition still faces the problem that only few datasets with mostly binary classifications tasks exist. The datasets are often small in terms of size and class variety. Moreover, the images are taken under restricted laboratory conditions. They use only one camera with fixed focal length and a specific acquisition setup concerning the relative pose of camera and objects as well as lighting conditions. Real-world data, in contrast, is strongly heterogeneous because of the big variety of building types, environmental conditions and image qualities depending on the hardware and the inspector. 
This opens up the questions: 
how do models trained on existing RCD datasets perform on real-world data? 
How can existing open-source knowledge be exploited best and which existing datasets are useful? 
	
\begin{figure}
	\begin{center}
		\includegraphics[width=1\textwidth]{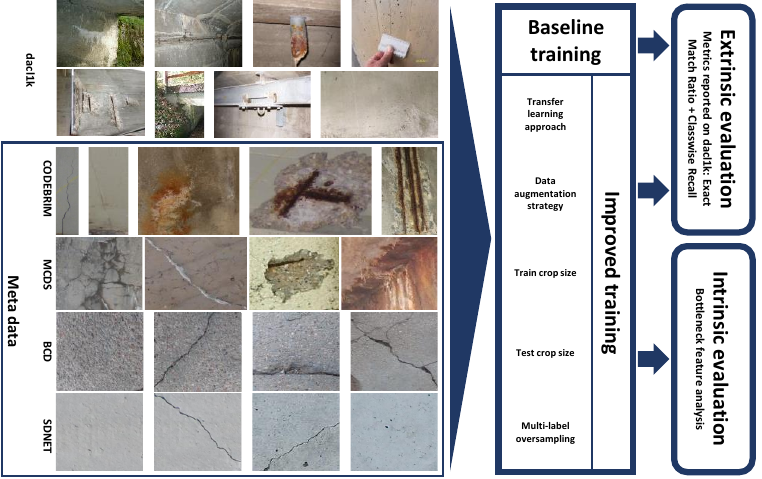}
		\caption{Example images from the four analysed open-source datasets and our dacl1k dataset. Firstly, we train models based on results of previous research (Baseline training) and then apply five improvements to the training process (Improved training). Secondly, models are evaluated on datasets (Extrinsic) and the best model is further analysed (Intrinsic).}
		\label{fig:MainScheme}
	\end{center}
\end{figure}

This work introduces the dacl1k (damage classification) dataset, the first RCD multi-label dataset that originates from real building inspections including 1,474 RCD images which were labeled by civil engineers inspired by German inspection standards. In order to train baseline models for dacl1k dataset, we conduct a proven transfer learning strategy, also called baseline training, and an improved training setting. 
For the improved setting, we examine five steps to leverage performance, choosing a different transfer learning approach, optimising the data augmentation, raising the training resolution, optimising the validation resolution and applying multi-label oversampling. 
Models resulting from both training strategies are evaluated on dacl1k. 
Thereby, we examine which open-source dataset combinations (we call them \textit{meta} datasets) are useful for our real-world defect images. 
The model showing best performance during extrinsic evaluation is subsequently evaluated intrinsically through a bottleneck feature analysis to further investigate the cause of its performance. 
Thereby, we use  dimensionality reduction techniques to identify, if the trained image representation are clustered according to data labels or data source. 
	
To summarise, we make the following contributions: 
(\textit{i}) dacl1k, the first RCD dataset for multi-label classification originating from real-world inspections, 
(\textit{ii}) models that result from computational expensive hyperparameter search and several improvements, 
(\textit{iii}) an extrinsic and intrinsic evaluation of the baselines. The dacl1k dataset, the meta datasets and the baselines will be made publicly available (see \url{https://github.com/phiyodr/building-inspection-toolkit}).

\section{Related Work}
\label{sec:relwork}
\textbf{Dataset compilations.} 
The Scene UNderstanding (SUN) database \cite{SUNdatabase} was generated by collecting images retrieved from multiple search engines that were proposed after seeking for terms that describe scenes, places, and environments. 
In the field of few-shot classification ten popular heterogeneous datasets, such as ImageNet~\cite{ILSVRC15} and MSCOCO~\cite{Lin2014} were combined to one meta dataset while two of them were used for validation \cite{triantafillou2020metadataset}. 
Others combined cross-domain and in-domain data to generate meta datasets for unlabeled, weakly-labeled or sparsely labeled target datasets \cite{ullah2023metaalbum, diagnostics13061068}. 
To the best of our knowledge, Kulkarni~et al.~\cite{CrackSeg9k} is the only work that combines previously available datasets, inter alia, from the RCD domain. They compile a semantic segmentation dataset, called CrackSeg9k, with 9,255 images of cracks from ten sub datasets on various surfaces. Before unifying the datasets, their individual problems (e.g. noise and distortion) are addressed by applying Image Processing.
SDNET~\cite{Dorafshan2018Dec}, which is used in the underlying work, is also part of CrackSeg9k for which no Image Processing was utilised.

\textbf{Transfer learning.}
Bukhsh~et al. \cite{Bukhsh2021} analysed combinations of six RCD (binary and multi-label) datasets from which four are used in our analysis (see Section~\ref{sec:data}).
They aimed to find the most valuable transfer learning dataset for each RCD dataset using a model initialised with and without weights from ImageNet.

The best performance for all datasets was obtained when using models initialised with ImageNet weights. 
On CODEBRIM~\cite{Mundt_2019_CVPR} the best performance was achieved when only weights from ImageNet were used, but no other RCD dataset. Thus, no additional training with an in-domain dataset was beneficial. 
MCDS~\cite{Huthwohl2019}, in contrast, benefited from training on the CODEBRIM dataset. 
The performance on each binary crack dataset, including SDNET~\cite{Dorafshan2018Dec} and BCD~\cite{Xu2019}, benefited from training with a different crack dataset. 
	
\textbf{Improvements.}
There are several levers that can be applied to enhance model performance, e.g. increasing or decreasing model capacity, regularising features or improving optimisation.
Many improvements regarding model capacity for datasets in the RCD domain were examined in previous work. 
In  \cite{bikit_icip} an extensive hyperparameter tuning for several state-of-the-art CNNs and transfer learning strategies were analysed. E.g., they examined different constellations of hidden layers in the classifier, optimiser types, learning rates and learning rate schedulers.
Yet, only basic image augmentation was used.
Unlike these approaches, we focus on the problem of regularisation by utilising basic image transformations, such as random horizontal and vertical flip, Random Erasing~\cite{zhong2017random} as well as automatic augmentation methods combined with a multi-label oversampling approach. 
There are several advanced augmentation algorithms (Random Erasing~\cite{zhong2017random}, AugMix~\cite{hendrycks2020augmix}, AutoAugment~\cite{cubuk2019autoaugment}, RandAugment~\cite{cubuk2019randaugment}, TrivialAugment~\cite{Muller_2021_ICCV}) available, which use a large set of image mutations and help to improve classification performance. Most of these methods provide a set of augmentations which are applied at a number of so-called "strength bins". In addition, the range of their augmentation strengths can be defined.
The final augmentation, Random Erasing, selects an arbitrary image region and overwrites its pixels with random values. 
	
Furthermore, current work \cite{Mundt_2019_CVPR}  suggests to increase the train crop size in the RCD domain. 
Recent work demonstrated that directly integrating multiple resolutions inside the network at train and test time raises performance, especially in category-level detection \cite{lin2017feature}. 
	
Increasing evaluation resolution compared to training resolution improved performance in previous work \cite{TouvronVDJ19}.
This may be beneficial for our models, although, we don't expect a notable disparity in the size of objects observed by the network between train and test phase since our image transformation, in both phases, includes no random resizing and cropping but resizing and centre-cropping.

\section{Datasets}
\label{sec:data}
For our experiments we use four open-source RCD datasets.
We call each of their combination ``meta dataset'' from which three versions exist.
Moreover, we introduce our real-world dataset dacl1k. 
This dataset is used for the evaluation of models trained on the meta data with respect to practical use.
Example images are shown in Figure~\ref{fig:MainScheme}.

\textbf{Open-source datasets.}
We use four open-source datasets which are relevant to us.
There are two binary crack datasets, \textbf{BCD}~\cite{Xu2019} and \textbf{SDNET}~\cite{Dorafshan2018Dec}. 
While BCD targets bridge cracks, SDNET includes images of cracks on walls, decks and pavement. 
Both are highly standardised datasets regarding hardware, object distance and camera angle which is orthogonal to the reference plane. 
Since SDNET has many incorrectly labeled data, we used a cleaned version \cite{rosch2022bikit}.

The largest and most realistic dataset in terms of damage appearance is \textbf{CODEBRIM}~\cite{Mundt_2019_CVPR}. 
The images were taken in a less standardised setting in comparison to BCD and SDNET. 
CODEBRIM and dacl1k share the same damage classes, which are very relevant for real-world inspections. 
There are two issues considering the practical transferability  of this dataset to real-world scenarios. 
Firstly, CODEBRIM is made up from image crops, thus, single images are split into rectangular patches depending on the maximum size of the defects. 
In addition, undamaged surface is extracted to act as background (\textit{No Damage}), leading to atypical image shapes. 
Long patches are often associated with cracks (see most left CODEBRIM image in Figure~\ref{fig:MainScheme}) and second most frequently with long shaped exposed reinforcement bars. 
Secondly, due to this cropping approach, some resulting patches become very small. 
The minimum image height and width is 22 and 40 pixels respectively. 

As the second dataset including multiple classes, we use an updated version~\cite{rosch2022bikit} of \textbf{MCDS}~\cite{Huthwohl2019}. 
From originally eight classes, the labels ``scaling'' and ``spalling'' are merged, since they show the same defect type and only differ with regard to the cause of the damage. 
Moreover, the class ``general'' is removed which summarises graffiti and vegetation. 
This is done due to the fact that ``general'' neither represents severe damage nor can the non-existence of this class in the other datasets be guaranteed. 
Consequently, the other multi-class datasets would have to be screened for general defects and labeled to avert false labels in the meta dataset compilations.

\begin{table}
	\begin{center}
		\small
		\tabcolsep=0.1cm
		\caption{Dataset statistics of the four open-source datasets as well as dacl1k, displaying the number of classes, samples, image size (height x width) and their meta affiliation.}
		\label{tab:DatsetOverview}
		\begin{tabular}{lcccccccc}
			\toprule
			Dataset  & Class & Samples  & Image size (min, median, max)             & meta2 & meta3 & meta4\\ \midrule
			CODEBRIM     & 6     & 7729   & (22, 359, 3638)$\times$(40, 826, 5997) & \cmark & \cmark  & \cmark\\
			MCDS     & 8     & 2597       & (24, 356, 2830)$\times$(47, 692, 4585)         & \cmark & \cmark  & \cmark\\
			BCD      & 2     & 6069       & (224, 224, 224)$\times$(224, 224, 224)         & $-$    & \cmark  & \cmark  \\
			SDNET    & 2     & 55449      & (256, 256, 256)$\times$(256, 256, 256)         &  $-$   & $-$     & \cmark \\ \midrule
			dacl1k   & 6     & 1474       & (245, 1024, 5152)$\times$(336, 1365, 6000)     & $-$    & $-$     & $-$ \\ \bottomrule
		\end{tabular}
	\end{center}
\end{table}

\textbf{dacl1k.} 
\begin{figure}[b]
	\begin{center}
		\includegraphics[width=1\textwidth]{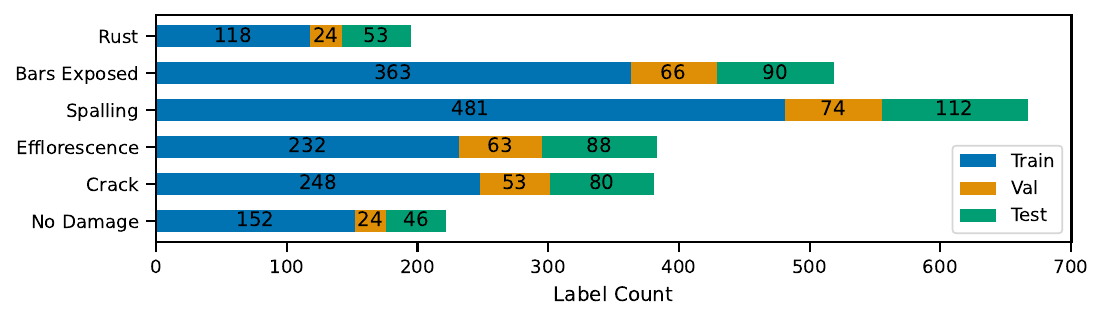}  
		\caption{Label distribution according to train, validation and test split in dacl1k.}
		\label{fig:Dacl1kSplits}
	\end{center}
\end{figure}
We release a novel multi-label classification dataset called dacl1k.
This dataset focuses on real-world inspection images in the RCD domain. 
While the heterogeneity of the dataset presents challenges for model training, it ensures that successful models have practical value in real-world scenarios. 
Our dataset includes five damage classes \textit{Crack}, \textit{Efflorescence}, \textit{Spalling}, \textit{Bars Exposed}, and \textit{Rust} and the label \textit{No Damage}. 
These classes are inherited from CODEBRIM with small changes to the original nomenclature. 
CODEBRIM is the most similar open-source dataset to dacl1k.
In contrast to CODEBRIM and MCDS, we supply uncropped images. 
Examples of dacl1k images are displayed in Figure~\ref{fig:MainScheme} where in the first two rows of the image tiles the following defects are shown: \textit{Cracks} with \textit{Efflorescence} (top-left), \textit{Crack} (top-right), \textit{Spalling} with \textit{Bars Exposed} and \textit{Rust} (bottom-left) and \textit{No Damage} (bottom-right). 
The dataset comprises a total of 1,474 images, each with a unique set of challenges including variations in camera types, poses, lighting conditions, and resolutions. 
The total number of labels amounts to 2,367. 
Our images derive from real inspections and were sourced from databases at authorities and engineering offices. 
We partitioned the dataset based on an equal label distribution into three subsets, with 67\% for training, 13\% for validation, and 20\% for testing (see Figure~\ref{fig:Dacl1kSplits}). 
Thus, dacl1k not only allows for testing but also training which is necessary due to lack of performance as described in Section~\ref{sec:experiments}. 

\textbf{Meta datasets.}
Our meta datasets are collections of aforementioned open-source data. 
The last three columns of Table~\ref{tab:DatsetOverview} indicate in which meta dataset the according open-source dataset is included. 
In order to build the meta datasets, all datasets are transformed to a six-class dataset according to dacl1k. 
The datasets are sequentially merged with the aim of assessing the impact of each additional data batch. 
We start with CODEBRIM because it is the most realistic and currently largest multi-label RCD dataset.
Subsequently, MCDS is added (\textit{meta2}) since it has the same classes, but being smaller in size.
Afterwards, we append BCD (\textit{meta3}) to increase the amount of crack images and healthy surface slightly.
Finally, we add SDNET to create \textit{meta4} which is the meta dataset with the strongest class imbalance due to the high amount of binary crack data.  

\section{Models}
\label{sec:experiments}
This section provides an overview of the trainings conducted during the development of the baselines for dacl1k. 
First, we examine a proven transfer learning approach named baseline training \cite{bikit_icip}. 
Second, we apply improvement steps to the training pipeline to further leverage performance. 

\textbf{Baseline training.}
\label{sec:DefTrain}
The settings of the default training are derived from previous work in RCD domain \cite{bikit_icip}. 
Here, three different CNN architectures and three transfer learning strategies were compared. 
Moreover, a computationally expensive hyperparameter search was conducted. 
In our baseline training we use their best MobileNetV3-Large~\cite{MobileNet} model.
Furthermore, their best transfer learning strategy called ``head then all'' (HTA) is applied. 
Here, in the first step the model base is frozen and only the classification head is trained. 
In the second step, all parameters can be updated. 
They applied basic image augmentation including resizing, cropping, random rotating and flipping. 
The training and validation crop size is 224$\times$224. 
The models are initialised with weights from ImageNet~\cite{ImageNet}.
Compared to \cite{bikit_icip}, we adjust the learning rate after a grid search while evaluating on the meta4 dataset.

\textbf{Improved training.}
\label{sec:ImprTrain}
To further improve model performance five additional setups are evaluated. 
Thereby, we focus on leveraging the performance of models fine-tuned on the most promising datasets, meta2+dacl1k and meta3+dacl1k as well as dacl1k itself. 
In order to improve the performance, we decided to examine another transfer learning approach applying different learning rates for model head and base (DHB)~\cite{bikit_icip, howard-ruder-2018-universal}. 

Furthermore, in initial experiments, we examined five different automatic augmentation methods: 
AugMix~\cite{hendrycks2020augmix}, AutoAugment~\cite{cubuk2019autoaugment}, RandAugment~\cite{cubuk2019randaugment}, TrivialAugment~\cite{Muller_2021_ICCV} (TA) and TA with a custom augmentation policy. 
The custom augmentation policy originates from our presumption that not all augmentations and their ranges inherited from the TA wide-custom augmentation space are beneficial to our data. 
Hence, we selected representative images from the dataset and evaluated each augmentation on the selection of data manually, based on subjective visual criteria. 
The custom policy neglects solarisation, posterisation and colourisation because they are considered as non-beneficial. 
Also, the minimum value of contrast augmentation is raised to 0.5 because the original range results in greying out the image completely. 
Before applying the augmentation method, random horizontal flip is applied and after the automatic augmentation method, Random Erasing~\cite{zhong2017random} is used.  
Supplementary, we execute a grid search for the ideal amount of magnitude bins in TrivAug. 
The search space is uniformly distributed in the interval 25 and 35. The best validation loss is obtained at a number of 34 magnitude bins.

In addition, training crop size is raised to 512$\times$512 in expectation of losing less image information due to resizing.
This may be beneficial to dacl1k with a relatively large resolution in comparison to the meta datasets (see Table~\ref{tab:DatsetOverview}). 
Another improvement step is the increase of the test resolution. 
Therefore, we test the models on different test crop sizes, starting from a size of 512$\times$512 and subsequently adding 16 pixels until 656. 

Finally, we tackle the problem of class imbalance by applying a multi-label oversampling strategy.
This is especially useful when fusing multi-class datasets with cracks-only datasets, such as SDNET, to correct the predominance of images showing cracks. 
The algorithm up-samples minority classes regarding the following steps:
(\textit{i}) calculate class counts in the current dataset and then calculate the standard deviation based of the count values,
(\textit{ii}) randomly draw an image from the dataset and update counts and standard deviation, 
(\textit{iii}) if the new standard deviation is reduced, this sample is added to the dataset, if not, another sample is drawn. 
This is repeated a fixed number of times. 

As in the default setting, both learning rates (for head and base) were adjusted based on a grid search.

\section{Evaluation}
In the following, we examine which dataset combination is the most valuable for practical use. 
Moreover, the best training settings regarding the improvement steps are analysed. 
In Section~\ref{sec:exrtinsic} the models from the baseline and improved training are evaluated. 
Section~\ref{sec:intrinsic} includes an intrinsic analysis of the model showing the best performance according to its extrinsic evaluation on dacl1k.

\subsection{Extrinsic evaluation}
\label{sec:exrtinsic}

We report the Exact Match Ratio (EMR) and classwise Recall to evaluate the models extrinsically. 
EMR is a challenging metric because all six label predictions of a given sample have to match the ground-truth. 
This metric is provided for the according source datasets (\textit{itself}) and for \textit{dacl1k} in Table~\ref{tab:MainTable}. 
The classwise Recall is relevant from a civil engineer's point of view because it displays how many defects are overlooked. 
We analyse results from models trained on CODEBRIM, meta2, meta3, meta4 and the combination with dacl1k datasets.

\begin{table*}
	\begin{center}
		\small
		\tabcolsep=0.1cm
		\caption{EMR on the model's source test set (itself) and on dacl1k with the according classwise Recall on dacl1k. All values in percent.}
		\label{tab:MainTable}
		\begin{tabular}{lcccccccc}
			\toprule
			\multicolumn{1}{l}{\multirow{2}{*}{Trained   on}}  & \multicolumn{2}{c}{EMR}          & \multicolumn{6}{c}{Recall on dacl1k}                                                                                                                                      \\ \cmidrule(lr){2-3} \cmidrule(lr){4-9}
			\multicolumn{1}{c}{}                                                      & itself & dacl1k         & \multicolumn{1}{r}{NoDam.} & \multicolumn{1}{r}{Crack} & \multicolumn{1}{r}{Effl.} & \multicolumn{1}{r}{Spall.} & \multicolumn{1}{r}{BExp.} & \multicolumn{1}{r}{Rust} \\ \midrule
			CODEBRIM                                                              & 70.57  & 16.89          & \textbf{73.91}             & 15.00                     & 34.09                     & 24.44                      & \textbf{9.43}                & \textbf{38.39}           \\
			meta2                                                                 & 70.41  & \textbf{17.35} & 63.04                      & 15.00                     & 37.50                     & \textbf{28.89}             & 7.55                         & 14.29                    \\
			meta3                                                                 & \textbf{81.52}  & \textbf{17.35} & \textbf{73.91}             & 12.50                     & \textbf{40.91}            & \textbf{28.89}             & 5.66                         & 16.07                    \\
			meta4                                                                 & 77.84  & 16.44          & 71.74                      & \textbf{21.25}            & 38.64                     & 27.78                      & 1.89                         & 7.14                     \\ \midrule
			dacl1k                                                                & 23.29  & 23.29         & \textbf{65.22}                      & 22.50                     & 43.18                     & 44.44                      & 35.85                        & 70.54                    \\
			meta2+dacl1k                                                         & 49.22  & \textbf{27.85}          & 63.04                      & 31.25                     & \textbf{53.41}                     & \textbf{61.11}                      & \textbf{41.51}                        & 74.11                    \\
			meta3+dacl1k                                                         & 75.22  & 27.40          & 60.87                      & \textbf{32.50}                     & 48.86                     & \textbf{61.11}                      & \textbf{41.51}                        & 68.75                    \\
			meta4+dacl1k                                                          & \textbf{76.81}  & 26.94          & 63.04                      & 31.25                     & 43.18                     & 53.33                      & 35.85                        & \textbf{75.89}                    \\ \bottomrule
		\end{tabular}
	\end{center}
\end{table*}

\textbf{Baselines results.}
The models trained on CODEBRIM and meta datasets show similar results, when evaluated on their own test split, compared to performances reported by others [REFS] ranging from 70.41\% to 81.52\% EMR (see upper part of Table \ref{tab:MainTable}). 
Weak performance is reported when the networks are evaluated on dacl1k (16.44\% to 17.35\%). 
The best results are obtained by meta2 and meta3.

The model trained on dacl1k achieves an EMR of 23.29\% (see lower part of Table \ref{tab:MainTable}), which is approximately a 6 percent points improvement over meta2 or meta3 . 
When meta datasets and dacl1k are combined, the performance on dacl1k raises. 
Models benefit from the additional amount of real-world data. 
The classwise Recall shows weak performance for models solely fine-tuned on meta data. 
Especially the classes \textit{Bars Exposed}, \textit{Rust}, and \textit{Crack} indicate that the domain shift between meta and target data is big. 
In other words, knowledge gathered from open-source data only, is in the current setting properly transferable to the real-world dataset. 
Like the analysis on EMR, only the combinations of dacl1k and meta data leverages Recalls compared to the dacl1k trained model. 
Considering all displayed metrics, training on meta2 together with dacl1k is the most promising approach, followed by meta3+dacl1k. Thus, we apply the improvements described in Section~\ref{sec:ImprTrain} to raise the model performance. 

\textbf{Improved results.}
Table~\ref{tab:ImprovTable} presents the best performances achieved after applying the improvements described in Section~\ref{sec:ImprTrain}. 
All models show better results when being trained at a crop size of 512 compared to 224. 
Additionally, all models benefit from an increase in test resolution. 
The best performance was achieved by training on meta3+dacl1k in combination with TrivialAugment, a test crop size of 624 and multi-label oversampling (32.42\%).
This is about a 1 percent point increase in comparison to training on dacl1k dataset only. 
However, training on meta2+dacl1k did not lead to a better performance compared to the model trained on dacl1k only. 
With respect to the best result from the default training (see Table~\ref{tab:MainTable}), EMR is increased by 4.57\%. 
Regarding the classwise Recall, it can be stated that -- apart from \textit{Crack} and \textit{Efflorescence} -- good performance is achieved. 
For comparison, previous work \cite{bikit_icip} reported on the balanced version of CODEBRIM an EMR of 74\% and the following classwise Recalls: 95\% (\textit{No Damage}), 88\% (\textit{Crack}), 76\% (\textit{Efflorescence}), 89\% (\textit{Spalling}), 93\% (\textit{Bars Exposed}), 85\% (\textit{Rust}). 

\begin{table}
	\begin{center}   
		\small
		\tabcolsep=0.1cm
		\caption{Best improvement setting regarding augmentation (Aug), test crop size (TestCS) and multi-label oversampling (MO). EMR is reported on the test split of the datasets the models were trained on (itself). EMR and classwise Recall is reported for dacl1k test set. }
		\label{tab:ImprovTable}
		\begin{tabular}{lcccccccccccc}
			\toprule
			\multirow{2}{*}{Trained on} & \multicolumn{3}{c}{Improvements} & \multicolumn{2}{c}{EMR}  & \multicolumn{6}{c}{Recall on dacl1k}            \\ \cmidrule(lr){2-4} \cmidrule(lr){5-6} \cmidrule(lr){7-12}
			& Aug     & TestCS   & MO      & itself     & dacl1k         & NoDam. & Crack & Effl. & Spall. & BExp. & Rust  \\ \midrule
			dacl1k                      & default  & 528    & $-$                   & 31.51          & 31.51     &  \textbf{73.91} & \textbf{42.50} & 52.27          & 68.89          & 56.60          & 68.75      \\
			meta2+dacl1k                & custom  & 560    & $-$                   & 61.14          & 31.51    & 65.22          & 31.25          & \textbf{54.55} & 65.56          & 56.60          & \textbf{74.11} \\
			meta3+dacl1k                & triv-aug & 624    & \cmark & \textbf{74.57} & \textbf{32.42}       & 67.39          & 36.25          & 50.00          & \textbf{73.33} & \textbf{60.38} & \textbf{74.11} \\ \bottomrule
		\end{tabular}
	\end{center}
\end{table}

\subsection{Intrinsic evaluation}
\label{sec:intrinsic}
Performance on our new dacl1k dataset is weak. 
Therefore, we want to analyse the capabilities of our best model intrinsically.
We want to understand if our model mainly gained information from the image content or the  dataset source.
In the desired setting the model should be able to differentiate between image content, which are the six labels, and not between image sources.

\textbf{Approach.}
Our implementation is as follows. 
We extract bottleneck features from our best model according to EMR on dacl1k (see Table~\ref{tab:ImprovTable}) and a model initialised with ImageNet weights for all five datasets.
We randomly keep 330 images per dataset to obtain an evenly distributed number of images per data source. 
Then, we run a non-linear dimensionality reduction from 960 to 2 dimensions using t-SNE \cite{vandermaaten2008TSNE}.
Here, we carefully select the t-SNE hyperparameters. 
We found a perplexity of 20, 5000 optimisation steps, and a learning rate of 200 useful.

\begin{figure*}
	\begin{center}
		\begin{subfigure}[b]{1\textwidth}
			\includegraphics[width=1\textwidth]{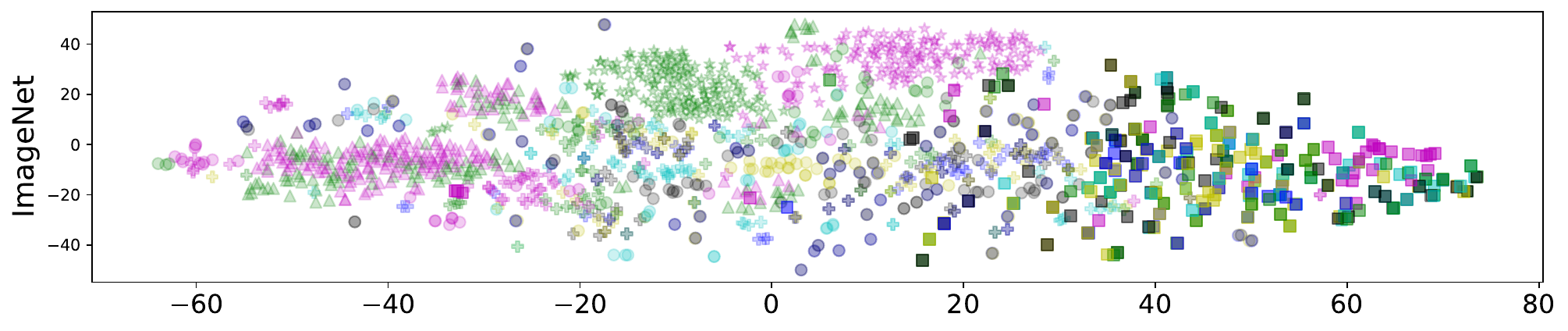}
		\end{subfigure}
		
		\begin{subfigure}[b]{1\textwidth}
			\includegraphics[width=1\textwidth]{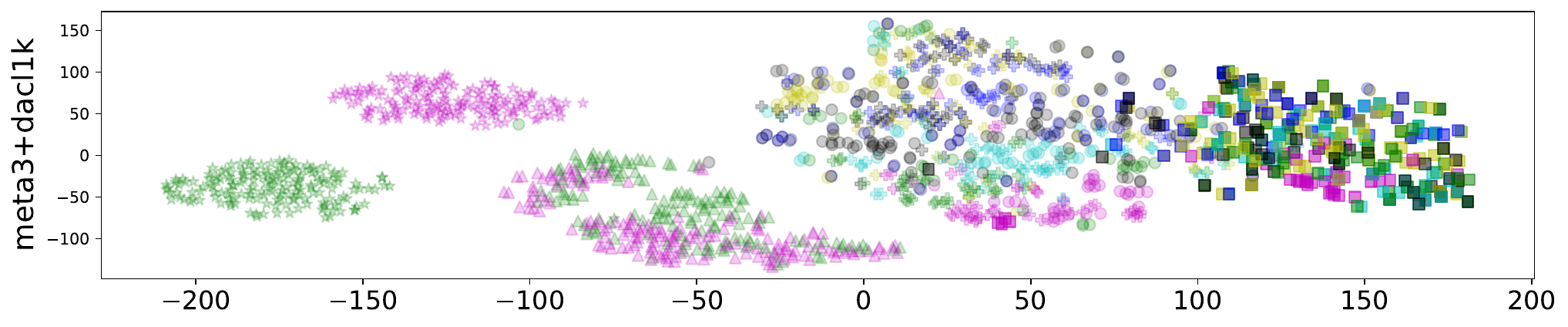}
		\end{subfigure}
		
		\begin{subfigure}[b]{1\textwidth}

			\includegraphics[width=1\textwidth]{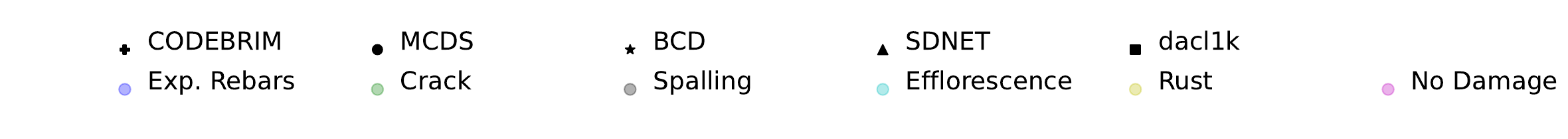}
		\end{subfigure}
		
		\caption{Clustering of images from all five datasets using bottleneck features after t-SNE dimensionality reduction. Features come from our best model trained on meta3+dacl1k (see Table \ref{tab:ImprovTable}). 
		}
		\label{fig:tsne} 
	\end{center}
\end{figure*}

\textbf{Results.}
In the visualisation of our best model in Figure~\ref{fig:tsne} we see clear clusters for the \textit{Crack} and \textit{No Damage} class of the BCD dataset.
Moreover, a dedicated cluster containing SDNET images only. 
Here, no clear distinction between the two classes is visible.
Right to the center, there is a mixed cluster for CODEBRIM and MCDS. No clear differentiation between the classes is recognisable.
On the far right one cluster for dacl1k datasets exists. 
Compared to the ImageNet visualisation on top, denser clusters are visible. 
This shows that the learned features have adapted to the underlying RCD domain.
However, the created clustering is still coarse and does not show clearly separated clusters according to damage types, but mainly relative to datasets.
To summarise, our best model mainly learned to differentiate between datasets and not between image content.

\subsection{Discussion}

Our work refers to the portability of features learned by open-source RCD data to our real-world dataset dacl1k.
During the development of our baselines, we face domain shifts regarding the underlying datasets. 
The first and largest shift is present between ImageNet and meta data because of the differences regarding their feature distribution and marginal distribution (6 vs. 1000 classes). 
The second and smaller shift appears as we combine the four different RCD datasets, which do not share the same marginal distribution by default (binary crack vs. multi-class). 
A third domain shift arises as we evaluate the models on real-world data while having pre-trained on ImageNet and trained -- apart from the meta-dacl1k combinations -- on meta data that is limited in terms of diversity \cite{Weiss2016Dec, SurveyTL}.
The performance displayed by the extrinsic evaluation (see Table~\ref{tab:MainTable} and Table~\ref{tab:ImprovTable}) indicates that models have difficulties to predict the samples in the dacl1k test set. 
Furthermore, the intrinsic analysis shows that the latent image representation adapts to the RCD domain but does not sufficiently learn features in order to clearly differentiate between damage classes. 
Another reason for the underlying performance can be shortcuts, or rather decision rules, learned from the source dataset hindering generalisation. These shortcuts can lead to the failure of models, especially, when they are tested on real-world data \cite{Geirhos_2020}.
Additional experimental results and discussions are presented in the supplementary material.

\section{Conclusion}
\label{sec:conclusion}
In this work, we presented dacl1k, a novel real-world dataset for the multi-label classification of defects occurring on massive bridges.
We investigated compilations of open-source datasets and improved the training process significantly over previous work. 
Selected models were evaluated extrinsically and intrinsically.
Results lead to the conclusion that in the RCD domain transferring knowledge from open-source data to real-world data shows weak performance. 
This also holds after having applied multiple improvement steps. 
The intrinsic evaluation underlines that the meta and dacl1k image representation strongly differs from each other.
Moreover, achieving a successful domain transfer between ImageNet, meta, and our dacl1k dataset requires further research. 
The applied improvement steps raised the performance significantly to an Exact Match Ratio of over 32\%. 
Yet, this is still insufficient for practical use in the digitised inspection framework. 
The classwise Recall shows that, apart from \textit{Crack} and \textit{Efflorescence}, 60 to 74\% of the real-world defects are recognised.

Our work can be a starting point to label further real-world data in a semi-automated fashion \cite{yu15lsun}. 
Here, our real-world dataset and corresponding models can be used to pre-label unseen images.  
In a subsequent step, labels can be assigned automatically when a certain prediction probability has been reached. 
If this is not achieved, the image must be annotated manually.
In addition, dacl1k enables estimating and comparing the usability of models for practical use. 
This is especially of great value for authorities which are confronted with products offering ``damage recognition through AI''. 
Our work acts as a benchmark for evaluating such applications.

\textbf{Acknowledgements.}
We are most grateful to Heiko Neumann and Johannes Kreutz for their insights and suggestions.
We thank the Bavarian Ministry of Economic Affairs for funding the MoBaP research project (IUK-1911-0004// IUK639/003) from which this work originates, the engineering offices and authorities for providing data. We gratefully acknowledge the computing time granted by the Institute for Distributed Intelligent Systems and provided on the GPU cluster Monacum One at the University of the Bundeswehr Munich.

\clearpage
\appendix
\section{Datasets}

\begin{figure}
	\footnotesize
	\begin{center}
		\begin{subfigure}[b]{01\textwidth}
			\includegraphics[height=0.135\textheight]{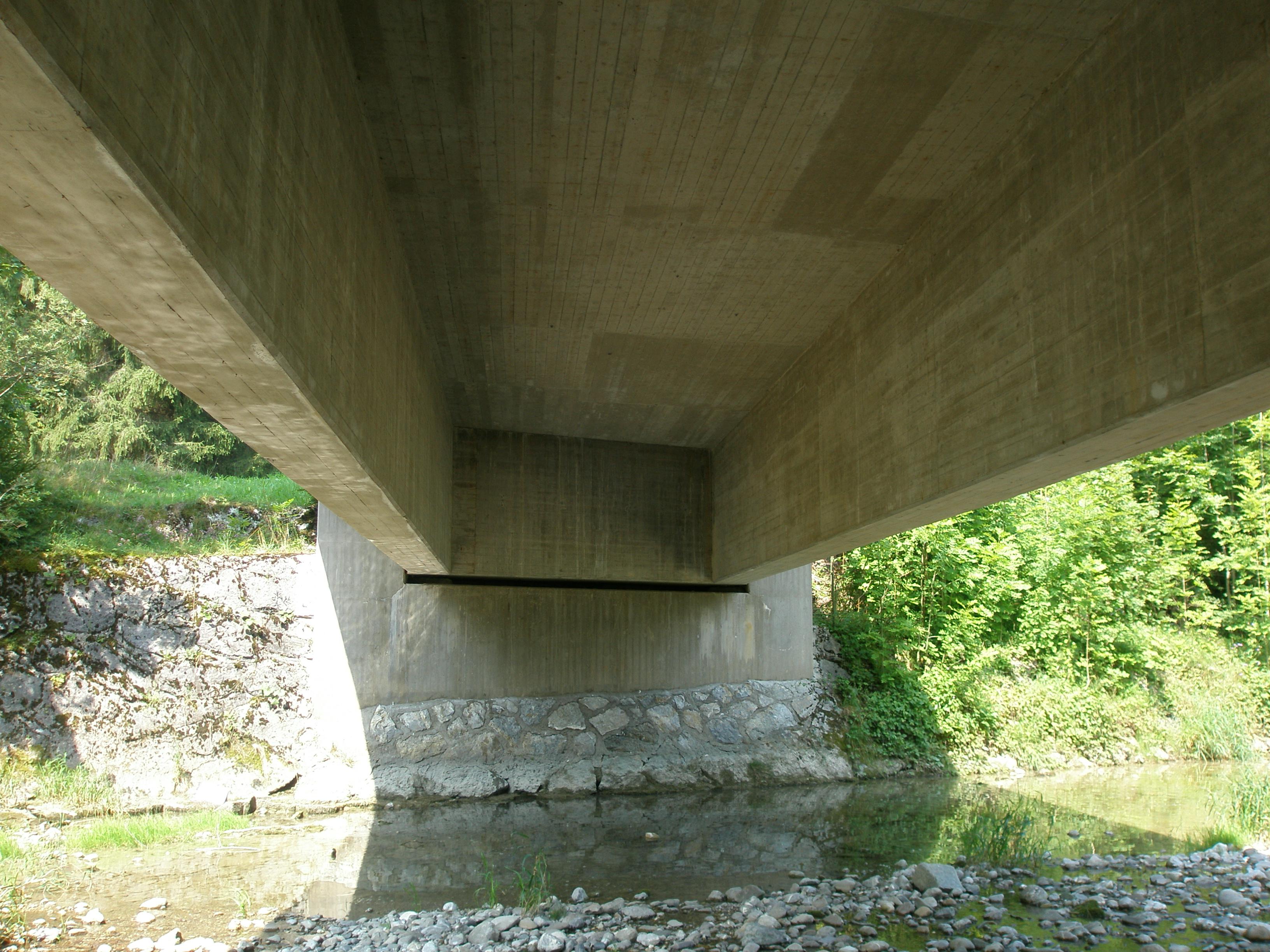}\hfill
			\includegraphics[height=0.135\textheight]{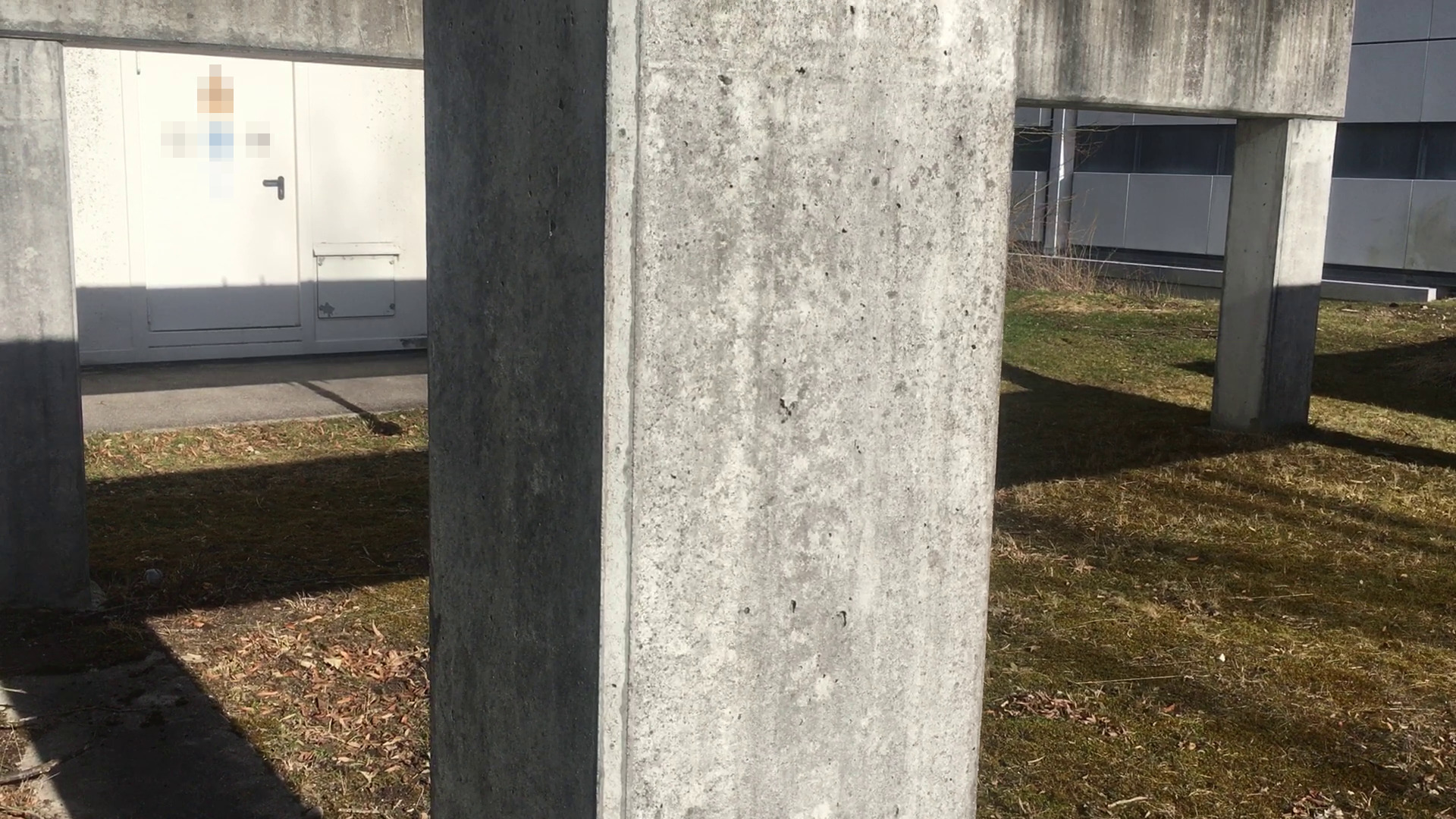}\hfill
			\includegraphics[height=0.135\textheight]{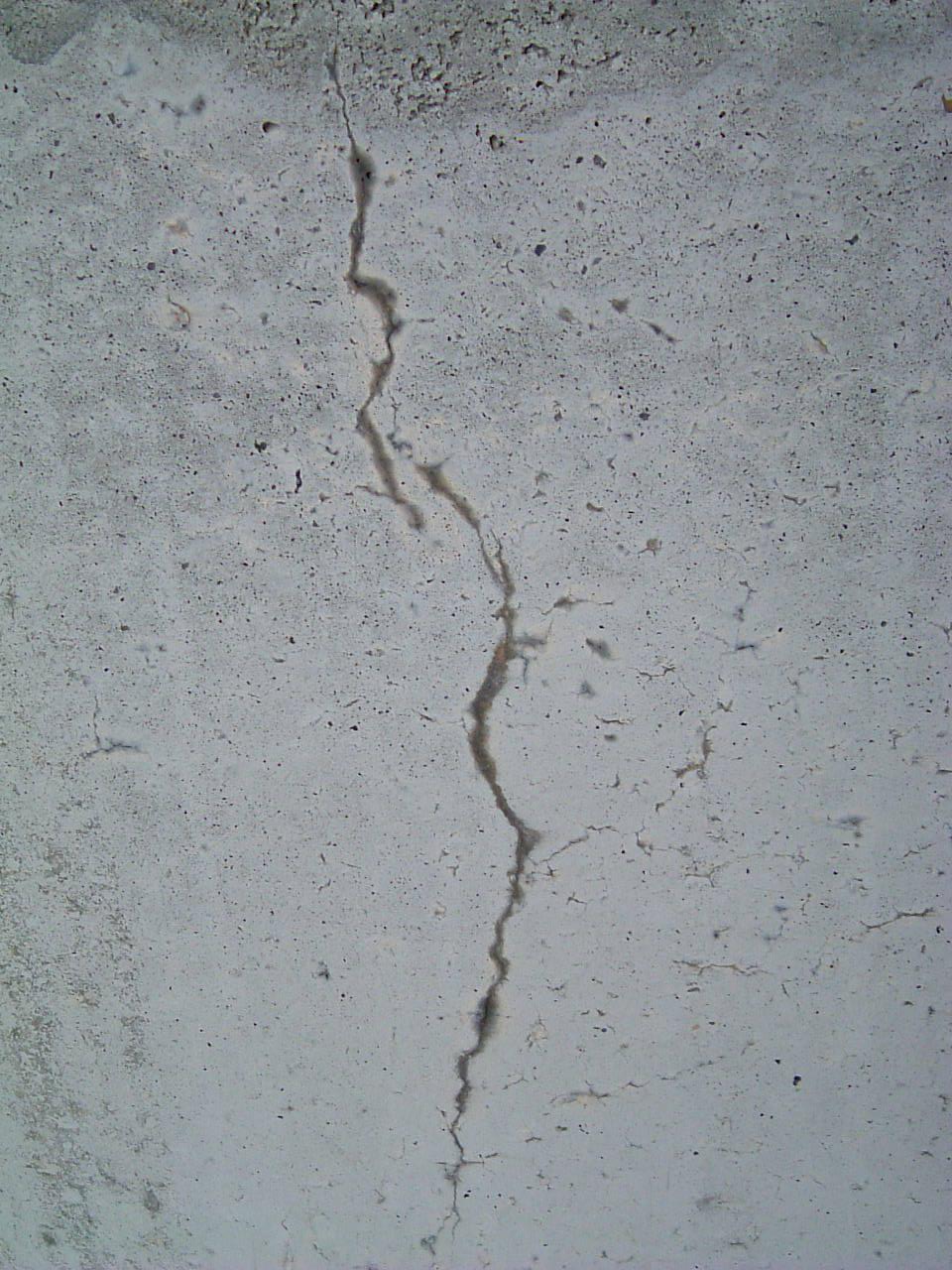}\hfill
			\includegraphics[height=0.135\textheight]{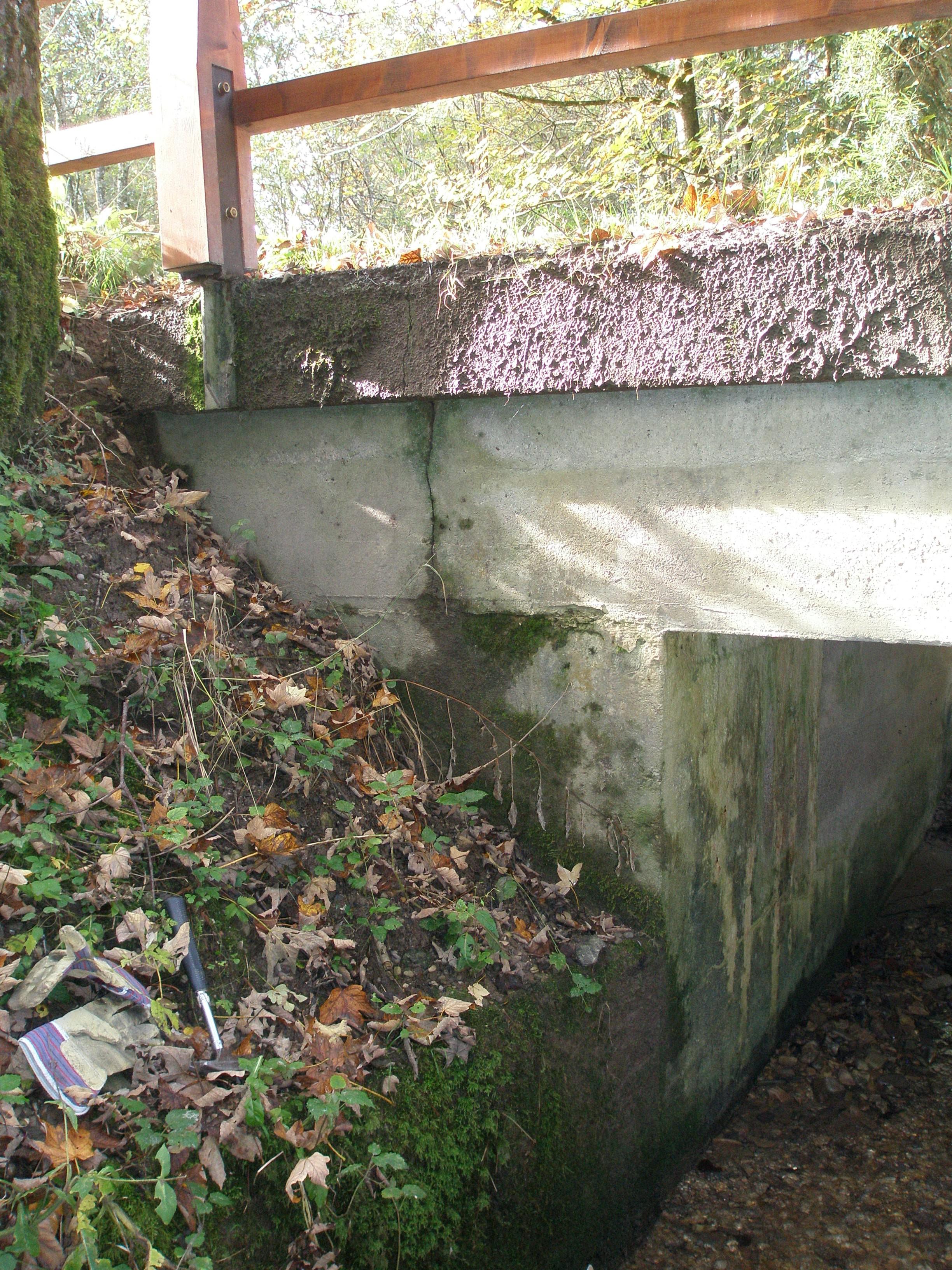}\hfill
			\includegraphics[height=0.135\textheight]{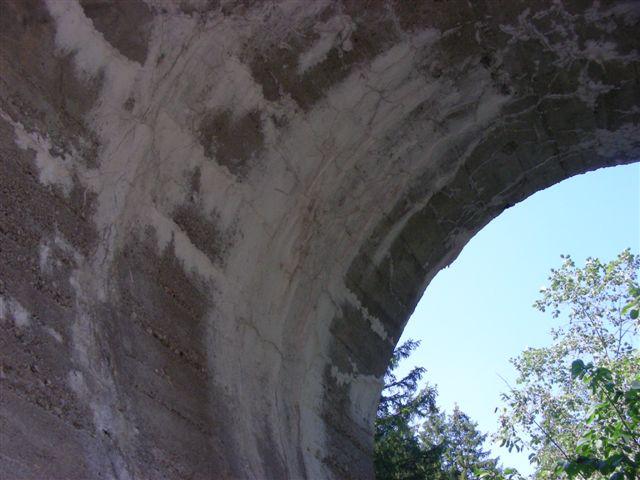}\hfill
			\includegraphics[height=0.135\textheight]{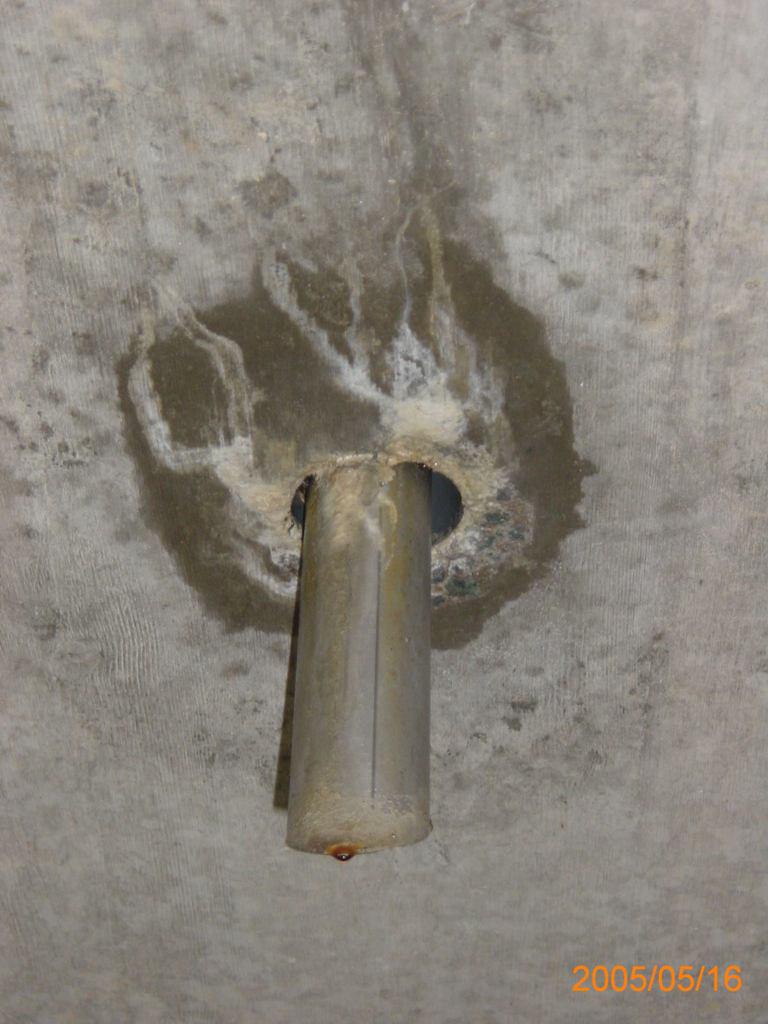}\hfill
			\includegraphics[height=0.135\textheight]{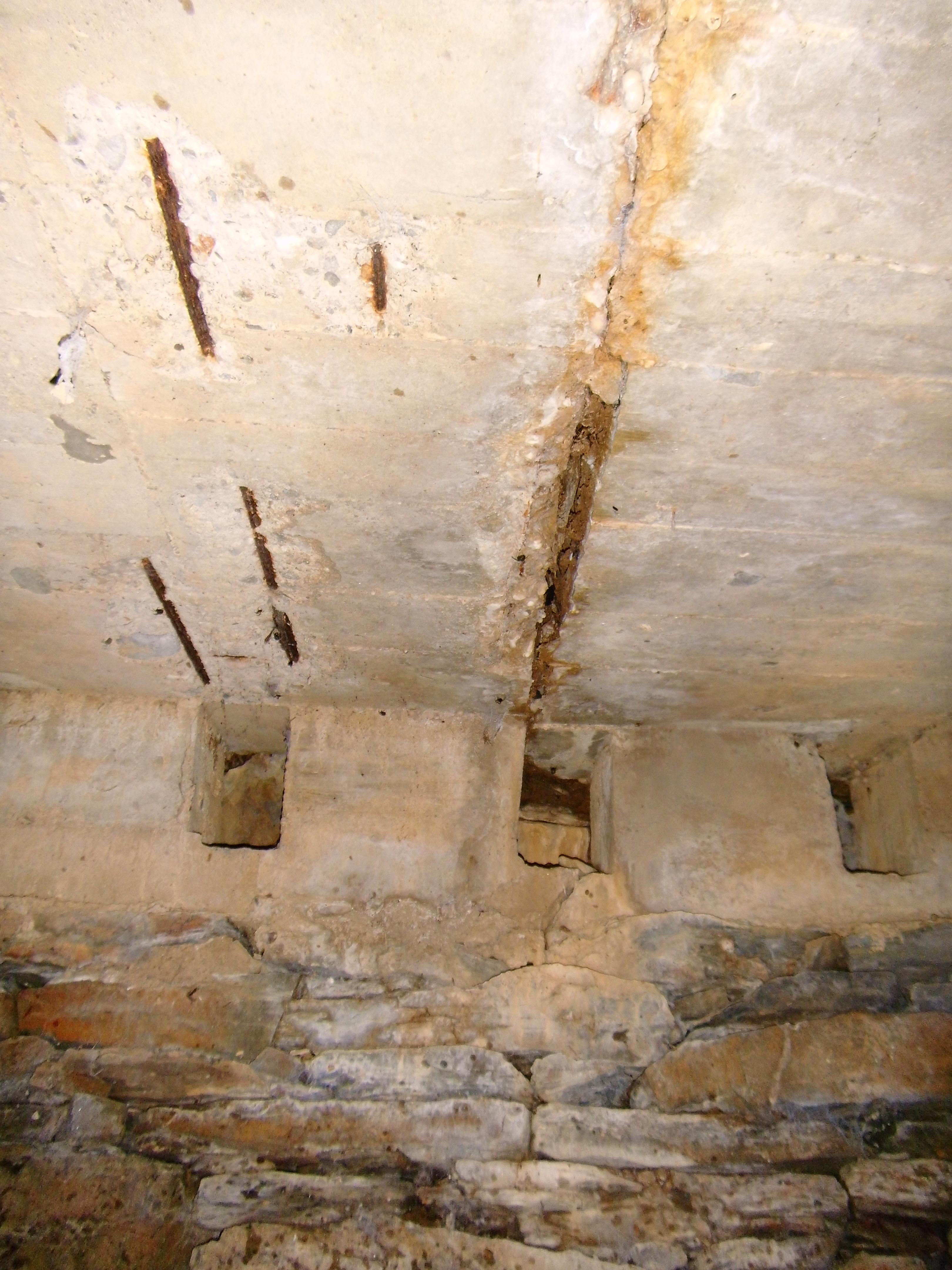}\hfill
			\includegraphics[height=0.135\textheight]{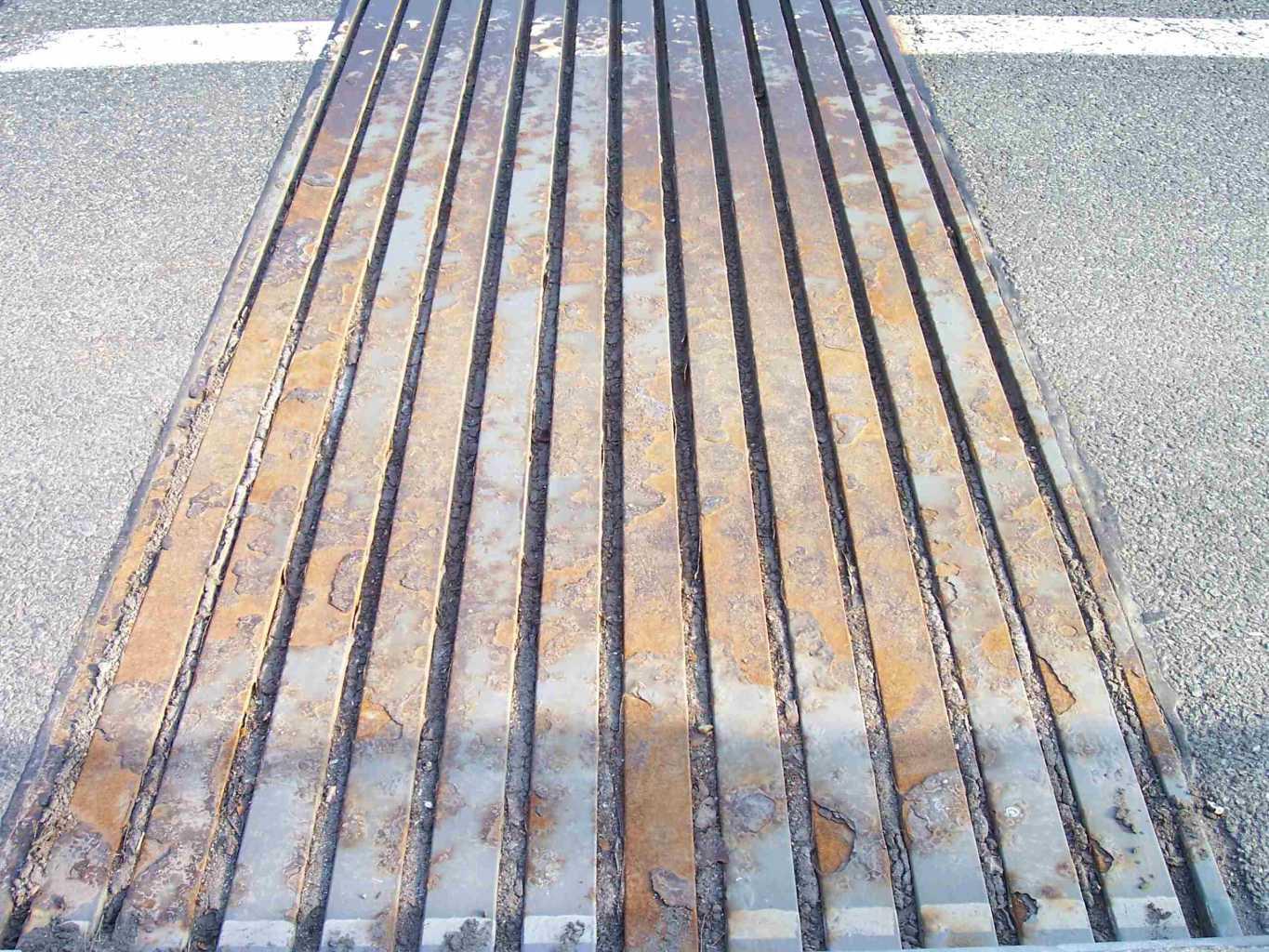}\hfill
			
			\caption{dacl1k. Top row, from left to right: \textit{No Damage}, \textit{No Damage}, \textit{Crack}, \textit{Crack}. Bottom row, from left to right: \textit{Efflorescence} with \textit{Crack}, \textit{Efflorescence} and \textit{Rust}, \textit{Spalling} with \textit{Bars Exposed} and \textit{Rust}, \textit{Rust}.}
			\label{fig:dacl1k} 
		\end{subfigure}
		\begin{subfigure}[b]{1\textwidth}
			\includegraphics[height=0.125\textheight]{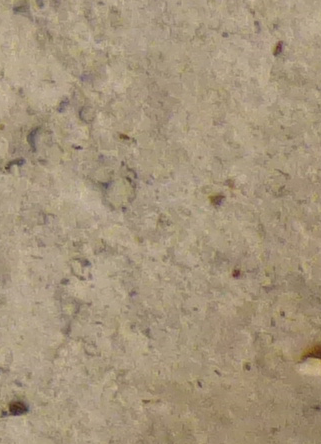}\hfill
			\includegraphics[height=0.125\textheight]{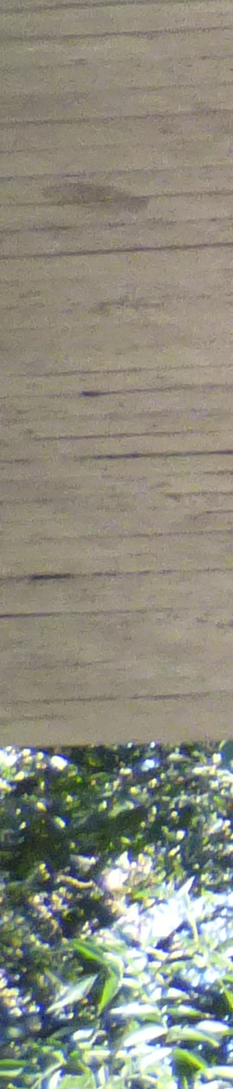}\hfill
			\includegraphics[height=0.125\textheight]{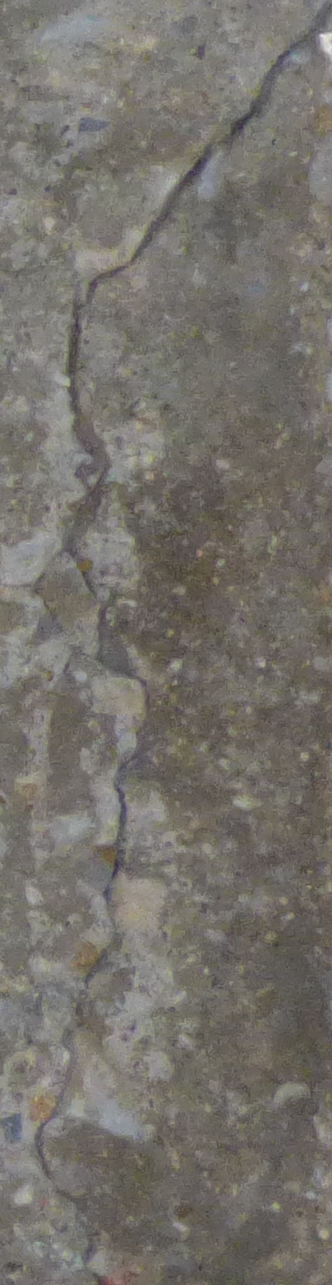}\hfill
			\includegraphics[height=0.125\textheight]{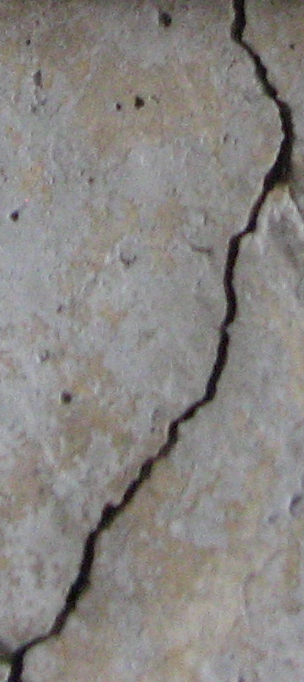}\hfill
			\includegraphics[height=0.125\textheight]{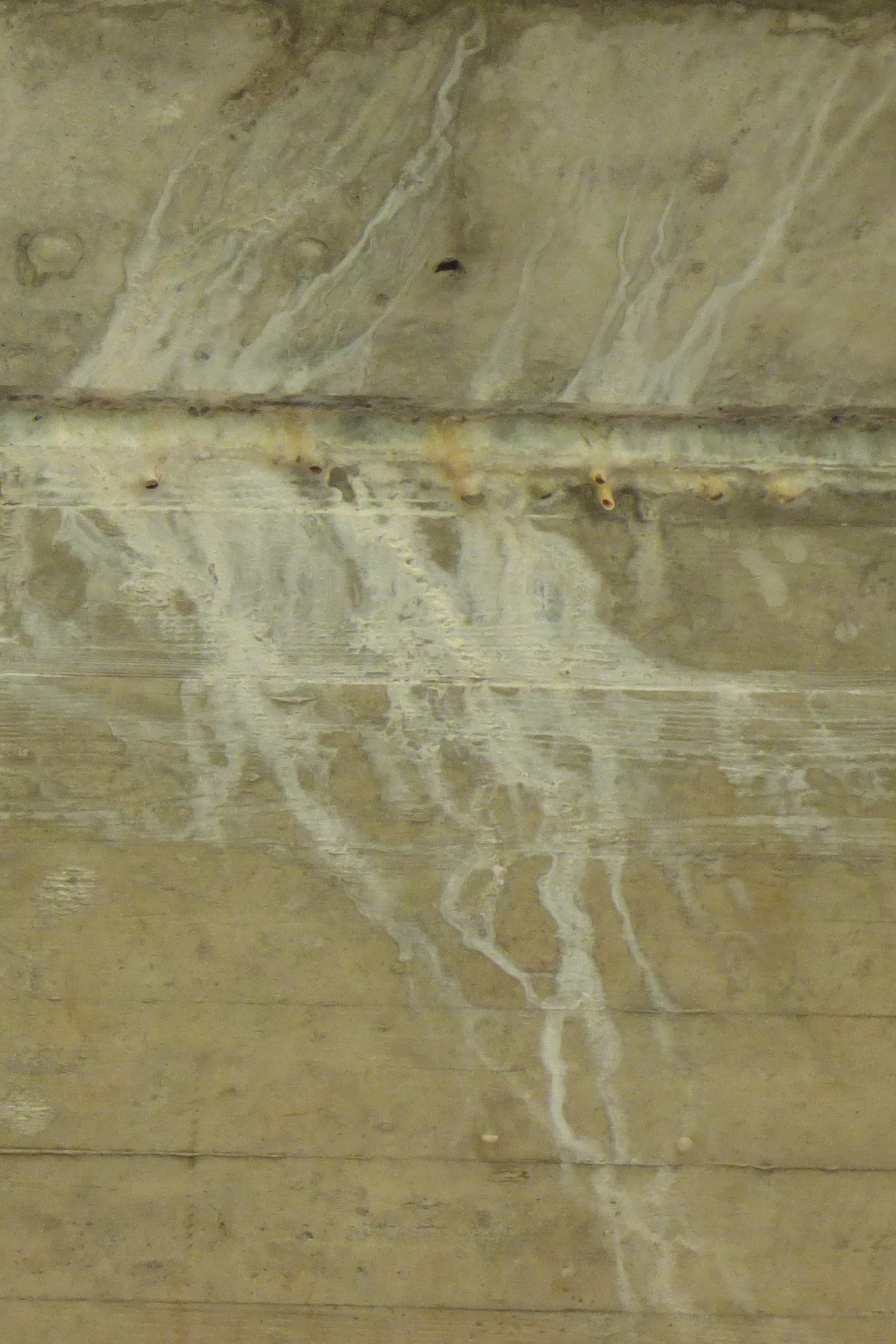}\hfill
			\includegraphics[height=0.125\textheight]{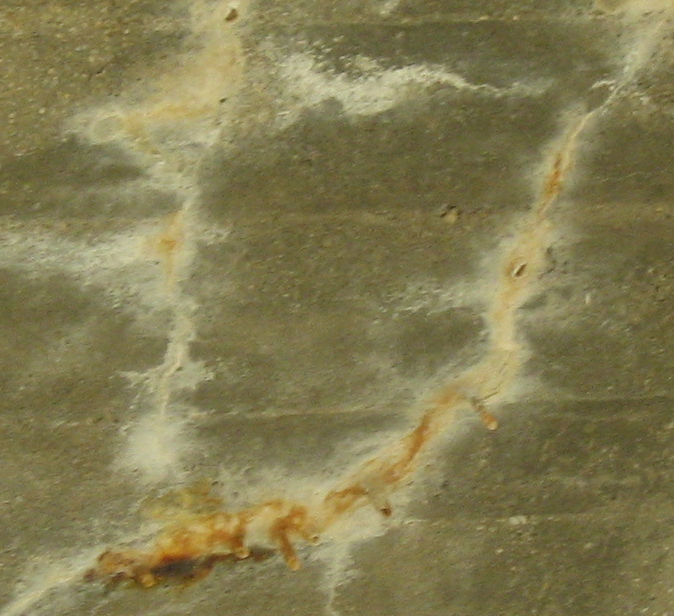}\hfill
			\includegraphics[height=0.125\textheight]{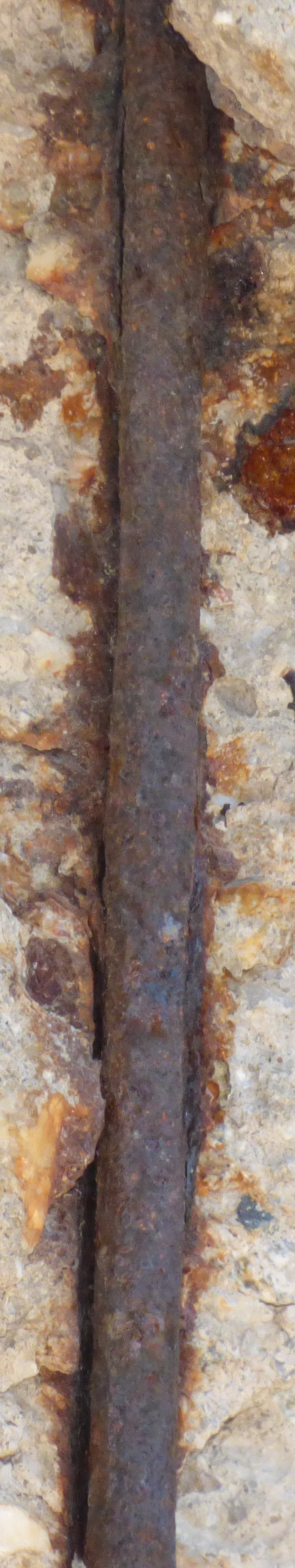}\hfill
			\includegraphics[height=0.125\textheight]{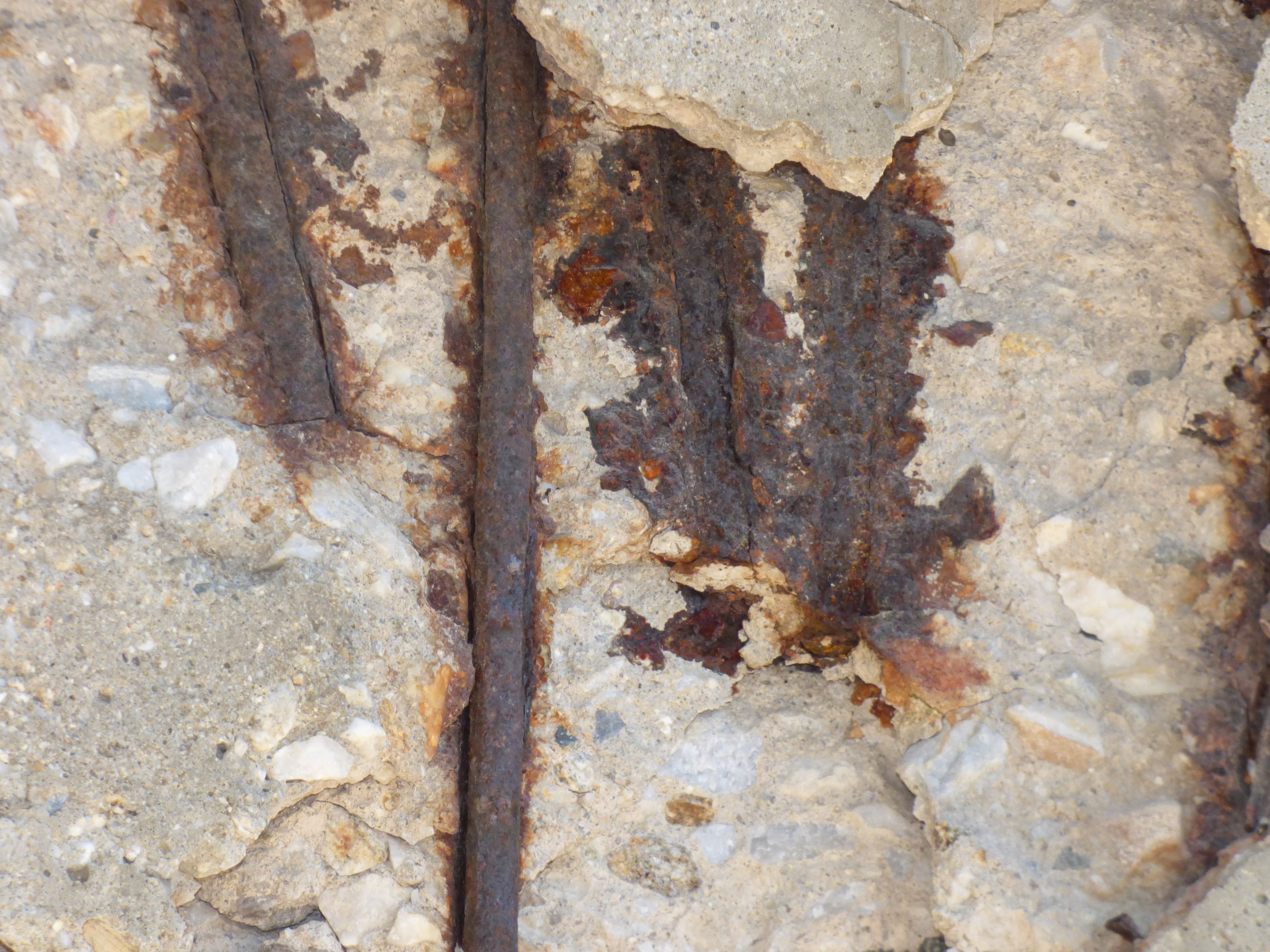}\hfill
			\caption{CODEBRIM. From left to right pairs of: ``Background'', ``Crack'', ``Efflorescence'', ``Spallation'' with ``ExposedBars'' and ``CorrosionStain''.}
			\label{fig:CODEBRIM} 
		\end{subfigure}
		\begin{subfigure}[b]{01\textwidth}
			\includegraphics[height=0.075\textheight]{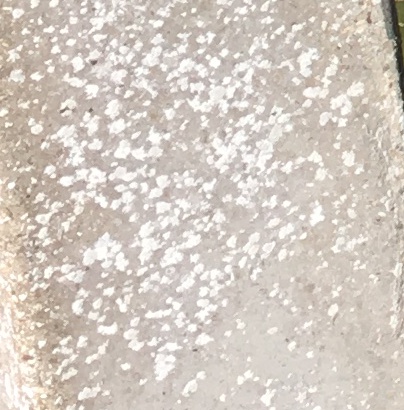}\hfill
			\includegraphics[height=0.075\textheight]{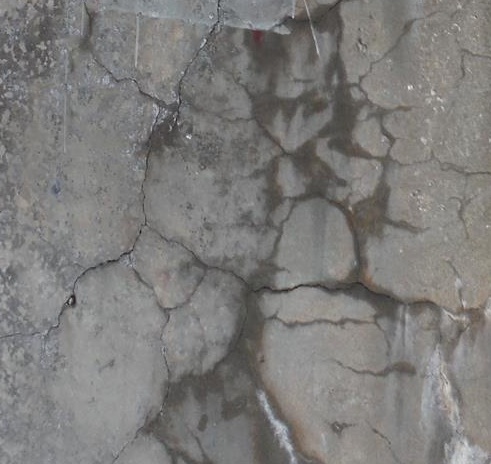}\hfill
			\includegraphics[height=0.075\textheight]{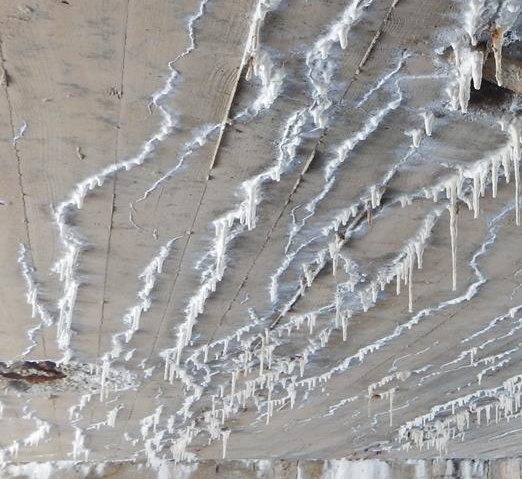}\hfill
			\includegraphics[height=0.075\textheight]{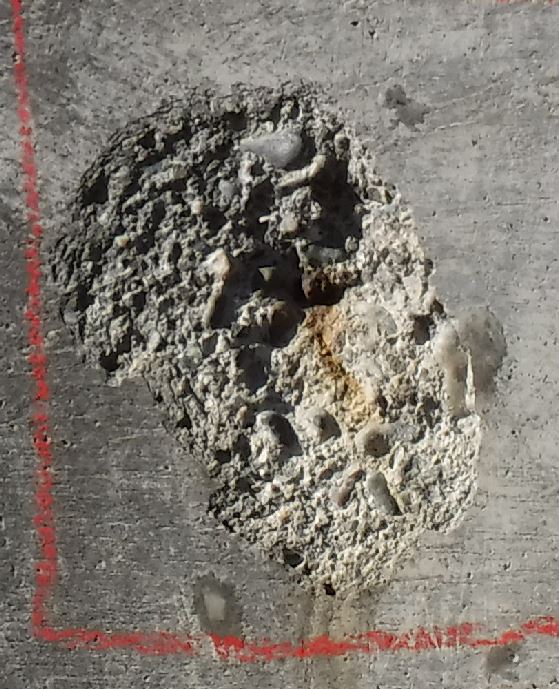}\hfill
			\includegraphics[height=0.075\textheight]{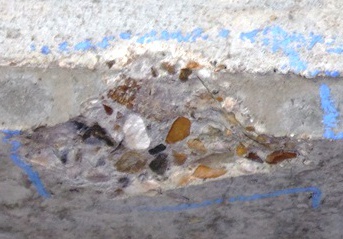}\hfill
			\includegraphics[height=0.075\textheight]{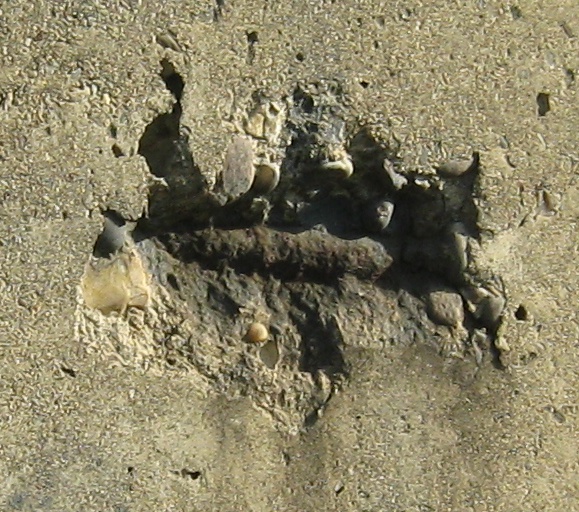}\hfill
			\includegraphics[height=0.075\textheight]{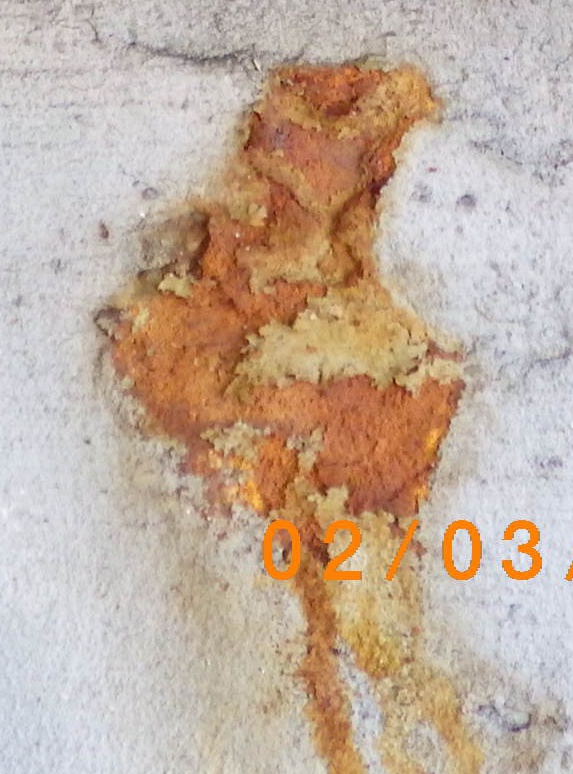}\hfill
			\includegraphics[height=0.075\textheight]{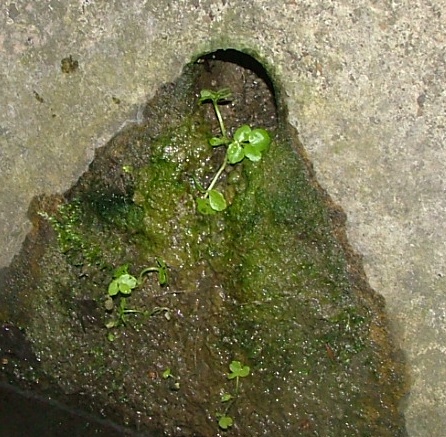}\hfill
			\caption{MCDS. One sample for each damage class. From Left to right: ``NoDefect'', ``Cracks'', ``Efflorescence'', ``Scaling'', ``Spalling'', ``Exposed Reinfor.'' without ``Rust Staining'', ``Rust Staining'', ``General''.}
			\label{fig:MCDS} 
		\end{subfigure}
		\begin{subfigure}[b]{01\textwidth}
			\includegraphics[height=0.1\textheight]{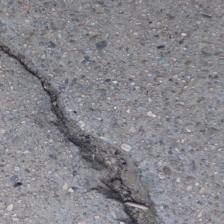}\hfill
			\includegraphics[height=0.1\textheight]{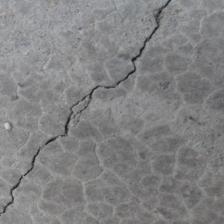}\hfill
			\includegraphics[height=0.1\textheight]{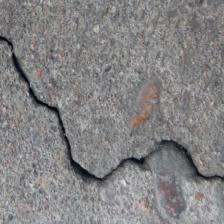}\hfill
			\includegraphics[height=0.1\textheight]{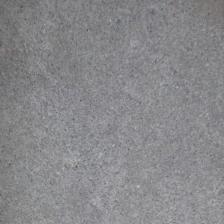}\hfill
			\includegraphics[height=0.1\textheight]{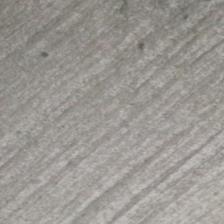}\hfill
			\includegraphics[height=0.1\textheight]{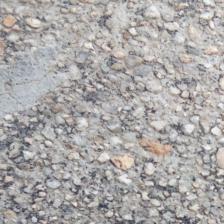}\hfill
			\caption{BCD. Three samples of the label ``Crack'' (left) and ``NoCrack'' (right).}
			\label{BCD} 
		\end{subfigure}
		\begin{subfigure}[b]{01\textwidth}
			\includegraphics[height=0.1\textheight]{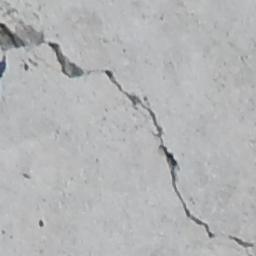}\hfill
			\includegraphics[height=0.1\textheight]{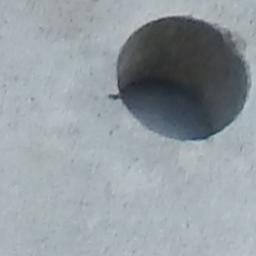}\hfill
			\includegraphics[height=0.1\textheight]{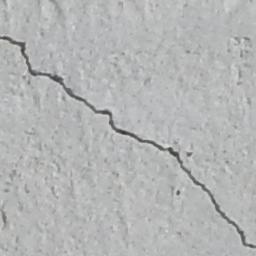}\hfill
			\includegraphics[height=0.1\textheight]{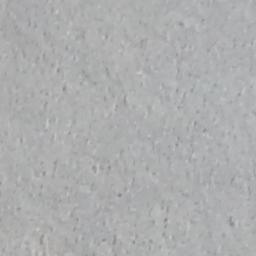}\hfill
			\includegraphics[height=0.1\textheight]{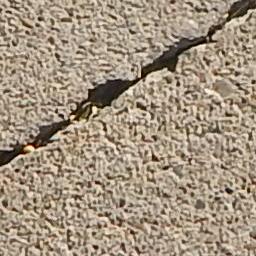}\hfill
			\includegraphics[height=0.1\textheight]{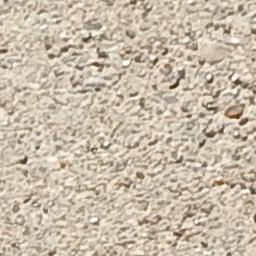}\hfill
			\caption{SDNET. Pairwise ``Crack'' (cracked) and ``Uncracked'' surfaces of different building parts, from left to right: wall, deck and pavement.}
			\label{fig:SDNET.} 
		\end{subfigure}
		\caption{Detailed view of all five datasets used in our experiments. The sub-captions include the according dataset's original damage names (see Table~\ref{tab:LabelNames}).}
		\label{fig:Images} 
	\end{center}
\end{figure}

During labeling and quality assessment for dacl1k we followed a two stage annotation process. First, civil engineering students labeled the real-world inspection images after having completed an initial training. Additionally, a detailed class and labeling guideline as well as a sample batch of labeled data was handed out. On the one hand, the class guideline clearly describes each damage class by naming the abbreviation, a detailed description of the visual appearance and the defect cause (see Table~\ref{tab:class_guide}). 
On the other hand, the labeling guideline points out the caveats based on previous labeling processes. Defects often overlap with each other, which often led to false negative labels because only the most obvious defects were recognised. The most common case for overlapping defects is \textit{Spalling} exposing the reinforcement (\textit{Bars Exposed}) which is covered by \textit{Rust}. Another common co-appearance is the combination of \textit{Efflorescence} and \textit{Crack}. The efflorescence stains complicate the detection of the subjacent cracks leading to many un-labeled cracks. Another error source is the presence of heavy weathering on surfaces that also show defects.    
After each initially submitted batch of data ($\approx$100 images) the labeling team continually received feedback and subsequently repaired their labels.

In a second quality assurance step, we appended samples showing \textit{No Damage} because the original label distribution showed an underrepresentation of healthy concrete surfaces. This is especially important for testing the models with regards to false positives.   
Furthermore, Figure~\ref{fig:Images} provides an overview of dacl1k's and the meta datasets' images in high resolution. The unified nomenclature from dacl1k and the meta datasets' original label names are displayed in Table~\ref{tab:LabelNames}. 

\begin{table}[h]
	\begin{center}
		\renewcommand{\arraystretch}{1.5}
		\caption{Descriptions of dacl1k's damage classes.}
		\label{tab:class_guide}
		\begin{tabular}{ll}
			\toprule
			\textbf{Damage}        & \textbf{Description}                                                                                                                                                   \\ \midrule
			No Damage     &  \parbox[t]{10cm}{No Damage label describes images that contain healthy concrete surface or irrelevant content.}                                                                                                                                                             \\ \midrule
			Crack         & \parbox[t]{10cm}{Cracks appear when the concrete's tensile strength is exceeded or during hardening, if the post-treatment or the concrete recipe was inadequate.}             \\ \midrule
			Efflorescence & \parbox[t]{10cm}{Efflorescence is usually whitish, or yellowish. It appears when salts (calcium, sodium, potassium) of the cement stone get dissolved. Cement stone is the hardened cement paste (cement and water) that binds the other concrete components, sand and gravel (aggregate), together. This is a frequently occurring defect emerging when the according building part is constantly in contact with running water.} \\ \midrule
			Spalling      & \parbox[t]{10cm}{Spalling can appear due to freeze thaw changes, corrosion of the subjacent reinforcement or impact, e.g. from cars that hit the structure.}  \\ \midrule
			Rust          & \parbox[t]{10cm}{Rust appears on metallic objects such as reinforcement and concrete. Rust that is visible on the concrete surface originates from neighbouring metallic parts or the subjacent reinforcement. Reinforcement can corrode as a result of loss of the alkaline protective layer provided by un-carbonated concrete. If the pH value drops due to the carbonation of the concrete, which is unavoidable over time, the reinforcement can oxidise.} \\ \midrule
			BarsExposed   & \parbox[t]{10cm}{Reinforcement that is visible due to the spalling of the overlying concrete, which is usually the consequence of corrosion of the reinforcement. Another cause for visible reinforcement are rockpockets which arise if the cement paste didn't fill all volume between the coarse aggregate. Rock pockets can follow, if the concrete's rheological properties or the compacting of the concrete was inadequate.}    \\ \bottomrule                                                          
		\end{tabular}
	\end{center}
\end{table}

\begin{table*}[!h]
	\begin{center}
		\caption{Our unified nomenclature in the first column and the original class names in BCD, SDNET, MCDS, and CODEBRIM. General damage class from MCDS is not considered.}
		\label{tab:LabelNames}
		\begin{tabular}{@{}lllll@{}}
			\toprule
			\textbf{meta4/dacl1k} & \textbf{BCD}  & \textbf{SDNET} &\textbf{MCDS} & \textbf{CODEBRIM}       \\ \midrule
			\textbf{NoDamage}      & NoCrack & Uncracked  & NoDefect             & Background     \\
			\textbf{Crack}        & Crack         & Crack    & Cracks                & Crack          \\
			\textbf{Efflorescence} & $\oldemptyset$   & $\oldemptyset$       & Efflorescence         & Efflorescence  \\
			\textbf{Spalling}     & $\oldemptyset$   & $\oldemptyset$       & Scaling; Spalling     & Spallation     \\
			\textbf{Rust}          & $\oldemptyset$     & $\oldemptyset$     & Rust Staining         & CorrosionStain \\
			\textbf{BarsExposed}   & $\oldemptyset$    & $\oldemptyset$      & Exposed Reinfor. & ExposedBars    \\
			$\oldemptyset$    & $\oldemptyset$   & $\oldemptyset$    & General               & $\oldemptyset$               \\ \bottomrule
		\end{tabular}
	\end{center}
\end{table*}

\section{Training settings}
\label{sec:trainset}
In the following, additional information regarding the training procedure is documented, concentrating on parameters deviating from \cite{bikit_icip}. 

{\textbf{Baseline training.}}
For the activation of the network's output layer, a Sigmoid layer is applied while using Binary Cross Entropy to compute the loss.  
For the first training step (HO), the learning rate is chosen according to a grid search, while evaluating on meta4 dataset that is the largest meta dataset. 
The learning rate is varied in the interval $1e^{-2}$ and $1e^{-4}$ which resulted in the best value at $5e^{-4}$. The second training step (HTA), which includes the training of all layers of the model, has a learning rate of $1e^{-5}$.
We make use of the Adam optimiser with weight decay \cite{loshchilov2019decoupled} together with a learning rate scheduler (cosine with warm-up). 
We trained the models for 100 epochs. The image pre-processing was held constant as follows: Resizing to 1.1*train resolution using bilinear interpolation and center-crop according to the chosen test crop size. 
We repeated each training with the same setting two times at different seeds. Finally, the best of the two models with respect to the maximum EMR is reported.

{\textbf{Improved training.}}
In contrast to the baseline training, for the improved training, 50 epochs were considered and the DHB approach was utilised. As in the default setting, both learning rates (for head and base) were adjusted based on a grid search. This led to a base learning rate of $1e^{-5}$ and a head learning rate of  $1e^{-3}$. Regarding the data augmentation, it is important to note that before applying the automatic augmentation method, random horizontal flip is applied with a probability of 50\%. 
After the augmentation method, Random Erasing~\cite{zhong2017random} is used with a probability of 10\%. The exact setting of the used TrivAug parameters are shown in Table~\ref{tab:CustAug}.

\section{Further extrinsic evaluation}
\begin{figure}[!h]
	\begin{center}
		\includegraphics[width=1\textwidth]{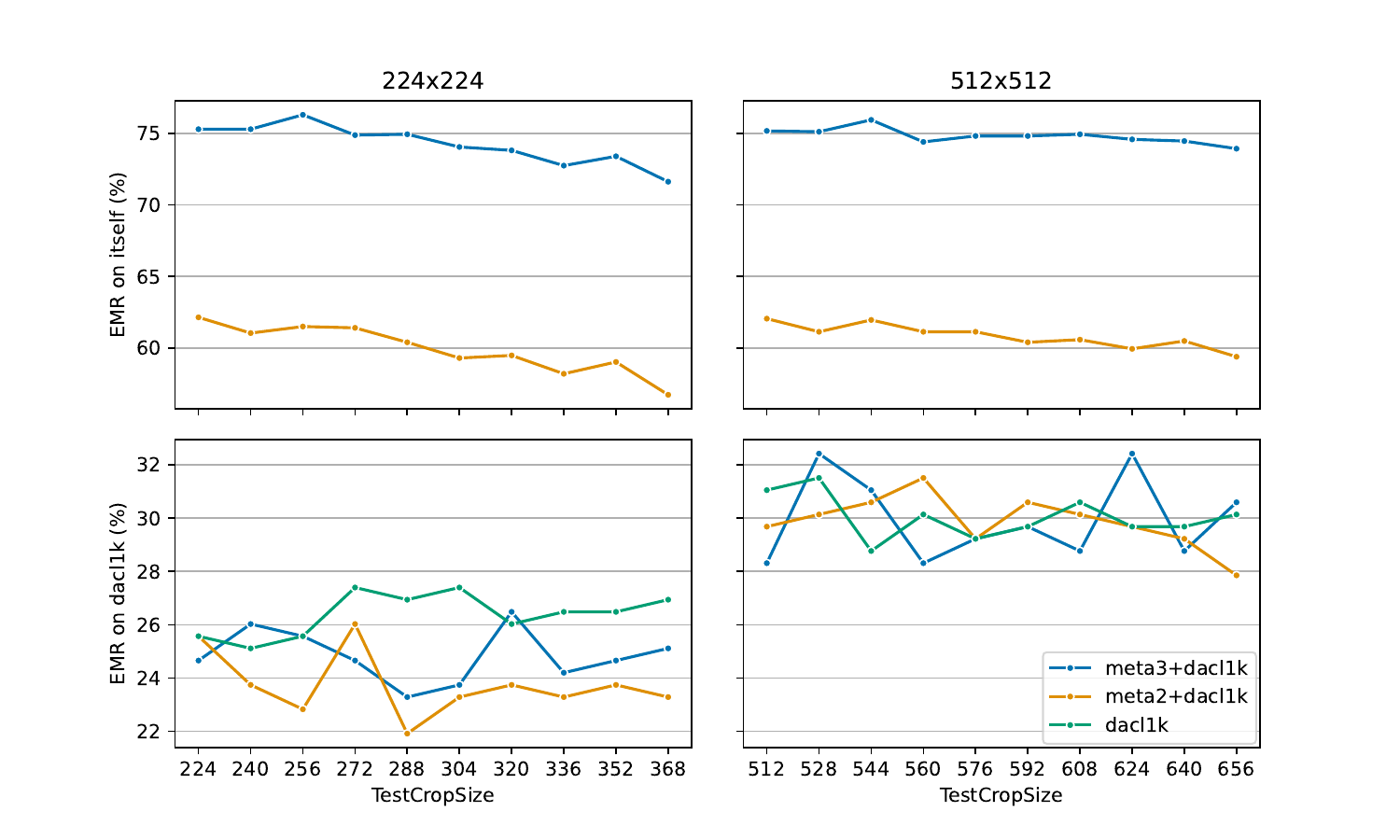}  
		\caption{EMR at varying test crop size on the according model's test split (itself) and on dacl1k. Three differently trained models are evaluated (dacl1k, meta2+dacl1k and meta3+dacl1k). Each model was trained in compliance with the best training setting (see Table 3 in the main paper for resolution 512x512 and Table~\ref{tab:FurtherExtrResults224} for 224x224). The charts share the y-axis row-wise and the x-axis column-wise.} 
		\label{fig:TestCropSize}
	\end{center}
\end{figure}

In the following, further extrinsic results are reported. The displayed models were trained on images with a resolution of 224x224 (see Table~\ref{tab:FurtherExtrResults224}) and 512x512 (see Table~\ref{tab:FurtherExtrResults512}). We only show results for models trained and tested with the same resolution in the tables. The investigations regarding the test resolution are displayed in Figure~\ref{fig:TestCropSize}. 

Comparing both tables makes clear that training on a resolution of 512x512 shows better results on dacl1k compared to training on 224x224. This can not be stated for models that are evaluated on their affiliated test set. Here, the EMR on itself is reduced by approximately 2\% for meta+dacl1k and meta3+dacl1k. Comparing the on-itself EMR, the dacl1k model is the only one profiting from the higher train resolution. With respect to the bigger average resolution of samples in dacl1k, this is reasonable. The best EMR on dacl1k is achieved by the model trained on meta2+dacl1k making use of custom data augmentation (26.03\%). The best performance of models trained on the 512x512 train crop size can be observed for the dacl1k model making use of no improvement step (31.05\%).    

\begin{table*}
	\begin{center}
		\scriptsize
		\tabcolsep=0.1cm
		\caption{Improvement setting regarding augmentation (Aug), and multi-label oversampling (MO) as well as test results on dacl1k (except EMR on itself) for models fine-tuned on three different datasets with a train resolution of 224x224. Underlined values represent the best results depending on the training dataset. Bold values represent the best overall result for the according metric.}
		\label{tab:FurtherExtrResults224}
		\begin{tabular}{lcccccccccc}
			\toprule
			\multirow{2}{*}{Trained on}                                              & \multicolumn{2}{c}{Improvements}                  & \multicolumn{2}{c}{EMR} & \multicolumn{6}{c}{Recall on dacl1k}            \\ \cmidrule(lr){2-3} \cmidrule(lr){4-5} \cmidrule(lr){6-11}
			& Aug                   & MO                    & itself     & dacl1k     & NoDam. & Crack & Effl. & Spall. & Bexp. & Rust  \\ \midrule
			\multirow{6}{*}{dacl1k}       & \multirow{2}{*}{default}  & $-$                   & 25.11      & 25.11      & 58.70  & 27.50 & 43.18 & 47.78  & 43.40 & 72.32 \\
			&                           & \cmark & 24.20      & 24.20      & 67.39  & 28.75 & \underline{51.14} & 44.44  & 60.38 & 73.21 \\ \cmidrule(lr){2-11}
			& \multirow{2}{*}{triv-aug} & $-$                   & 24.66      & 24.66      & 60.87  & 26.25 & 38.64 & \underline{51.11}  & 49.06 & 68.75 \\
			&                           & \cmark & 25.11      & 25.11      & \underline{\textbf{71.74}}  & 28.75 & 44.32 & 43.33  & \underline{\textbf{69.81}} & \underline{\textbf{75.89}} \\ \cmidrule(lr){2-11}
			& \multirow{2}{*}{custom1}  & $-$                   & 23.74      & 23.74      & 63.04  & 27.50 & 43.18 & 43.33  & 50.94 & 70.54 \\
			&                           & \cmark & \underline{25.57}      & \underline{25.57}      & 69.57  & \underline{37.50} & 50.00 & 43.33  & 71.70 & 67.86 \\ \midrule
			\multirow{6}{*}{meta2+dacl1k} & \multirow{2}{*}{default}  & $-$                   & 62.42      & 23.29      & 56.52  & \underline{31.25} & 55.68 & \underline{58.89}  & \underline{50.94} & 54.46 \\
			&                           & \cmark & 63.34      & 21.92      & 54.35  & 21.25 & \underline{72.73} & 55.56  & 49.06 & \underline{67.86} \\ \cmidrule(lr){2-11}
			& \multirow{2}{*}{triv-aug} & $-$                   & 64.80      & 25.57      & 58.70  & 30.00 & 56.82 & 50.00  & \underline{50.94} & 63.39 \\
			&                           & \cmark & \underline{65.54}      & 24.66      & \underline{65.22}  & 22.50 & 52.27 & 52.22  & 47.17 & 66.96 \\ \cmidrule(lr){2-11}
			& \multirow{2}{*}{custom}   & $-$                   & 64.71      & \underline{\textbf{26.03}}      & 60.87  & 25.00 & 54.55 & 52.22  & 49.06 & 62.50 \\
			&                           & \cmark & 64.53      & 24.20      & 63.04  & 32.50 & 55.68 & 44.44  & \underline{50.94} & 61.61 \\ \midrule
			\multirow{6}{*}{meta3+dacl1k} & \multirow{2}{*}{default}  & $-$                   & 75.75      & 22.83      & 45.65  & 21.25 & 68.18 & 45.56  & 30.19 & \underline{75.00} \\
			&                           & \cmark & 75.64      & 19.63      & 32.61  & 27.50 & 57.95 & 50.00  & 28.30 & 59.82 \\ \cmidrule(lr){2-11}
			& \multirow{2}{*}{triv-aug} & $-$                   & 77.17      & \underline{25.11}      & \underline{63.04}  & 22.50 & 46.59 & \underline{\textbf{60.00}}  & 33.96 & 63.39 \\
			&                           & \cmark & 77.41      & 23.29      & 50.00  & 32.50 & \underline{\textbf{76.14}} & 36.67  & 47.17 & 61.61 \\ \cmidrule(lr){2-11}
			& \multirow{2}{*}{custom1}  & $-$                   & 76.64      & 24.66      & 58.70  & \underline{\textbf{36.25}} & 56.82 & 54.44  & 39.62 & 69.64 \\
			&                           & \cmark & \underline{\textbf{78.06}}      & 22.83      & 56.52  & 28.75 & 63.64 & 40.00  & \underline{54.72} & 61.61 \\ 
			
			\bottomrule
		\end{tabular}
	\end{center}
\end{table*}

\begin{table*}
	\begin{center}
		\scriptsize
		\tabcolsep=0.1cm
		\caption{Improvement setting regarding augmentation (Aug), and multi-label oversampling (MO) as well as test results on dacl1k (except EMR on itself) for models fine-tuned on three different datasets with a train resolution of 512x512. Underlined values represent the best results depending on the training dataset. Bold values represent the best overall result for the according metric.}
		\label{tab:FurtherExtrResults512}
		\begin{tabular}{lcccccccccc}
			\toprule
			\multirow{2}{*}{Trained on}                                              & \multicolumn{2}{c}{Improvements}                  & \multicolumn{2}{c}{EMR} & \multicolumn{6}{c}{Recall on dacl1k}            \\ \cmidrule(lr){2-3} \cmidrule(lr){4-5} \cmidrule(lr){6-11}
			& Aug                   & MO                    & itself     & dacl1k     & NoDam. & Crack & Effl. & Spall. & Bexp. & Rust  \\ \midrule
			\multirow{6}{*}{dacl1k}       & \multirow{2}{*}{default}  & $-$                   & \underline{31.05}      & \underline{\textbf{31.05}}        & 69.57  & 40.00 & 51.14 & \underline{71.11}  & 56.60 & 68.75 \\
			&                           & \cmark & 27.40      & 27.40        & \underline{\textbf{71.74}}  & 25.00 & 43.18 & 48.89  & 45.28 & \underline{78.57} \\ \cmidrule(lr){2-11}
			& \multirow{2}{*}{triv-aug} & $-$                   & 30.14      & 30.14        & 69.57  & 32.50 & 45.45 & 52.22  & 47.17 & 74.11 \\
			&                           & \cmark & 28.77      & 28.77        & \underline{\textbf{71.74}}  & 35.00 & \underline{53.41} & 43.33  & 49.06 & 68.75 \\ \cmidrule(lr){2-11}
			& \multirow{2}{*}{custom1}  & $-$                   & 23.74      & 23.74        & 56.52  & 32.50 & 39.77 & 46.67  & \underline{60.38} & 74.11 \\
			&                           & \cmark & 28.31      & 28.31        & 67.39  & \underline{\textbf{42.50}} & 50.00 & 38.89  & 56.60 & 65.18 \\ \midrule
			\multirow{6}{*}{meta2+dacl1k} & \multirow{2}{*}{default}  & $-$                   & 61.96      & 28.31      & 56.52  & \underline{38.75} & 55.68 & 58.89  & 54.72 & 69.64 \\
			&                           & \cmark & 62.24      & 24.20      & 52.17  & 35.00 & 52.27 & 40.00  & 50.94 & \underline{75.00} \\ \cmidrule(lr){2-11}
			& \multirow{2}{*}{triv-aug} & $-$                   & \underline{63.61}      & 28.77      & \underline{67.39}  & 35.00 & 51.14 & 65.56  & \underline{62.26} & 70.54 \\ 
			&                           & \cmark & 63.34      & 29.22      & 65.22  & 30.00 & \underline{\textbf{60.23}} & 55.56  & 60.38 & 72.32 \\ \cmidrule(lr){2-11}
			& \multirow{2}{*}{custom}   & $-$                   & 63.43      & \underline{29.68}      & 63.04  & 33.75 & 59.09 & \underline{71.11}  & 60.38 & 72.32 \\
			&                           & \cmark & 63.06      & 26.48      & 63.04  & 28.75 & 57.95 & 46.67  & 58.49 & 73.21 \\ \midrule
			\multirow{6}{*}{meta3+dacl1k} & \multirow{2}{*}{default}  & $-$                   & 74.51      & 24.66      & 54.35  & 25.00 & 44.32 & 64.44  & 47.17 & \underline{\textbf{79.46}} \\
			&                           & \cmark & 75.16      & 25.57      & 45.65  & 31.25 & 54.55 & 68.89  & \underline{\textbf{66.04}} & 67.86 \\ \cmidrule(lr){2-11}
			& \multirow{2}{*}{triv-aug} & $-$                   & 75.93      & \underline{28.77}      & 58.70  & \underline{36.25} & 51.14 & \underline{\textbf{74.44}}  & 52.83 & 73.21 \\
			&                           & \cmark & 75.93      & 28.31      & \underline{65.22}  & 30.00 & 52.27 & 67.78  & 60.38 & 71.43 \\ \cmidrule(lr){2-11}
			& \multirow{2}{*}{custom1}  & $-$                   & 76.17      & 28.31      & 56.52  & 32.50 & 46.59 & 70.00  & 49.06 & 72.32 \\
			&                           & \cmark & \underline{\textbf{76.40}}      & 24.20      & 43.48  & 30.00 & \underline{\textbf{60.23}} & 63.33  & 52.83 & 62.50 \\ 
			
			\bottomrule
		\end{tabular}
	\end{center}
\end{table*}

\begin{table}
	\begin{center}
		\footnotesize
		\tabcolsep=0.1cm
		\caption{Our custom augmentation pipeline based on Trivial Augment Wide~\cite{Muller_2021_ICCV} with manipulated range of contrast in bold.}
		\label{tab:CustAug}
		\begin{tabular}{lll}
			\hline
			& Augmentation           & Range/Probability      \\ \hline
			Basic augmentation                   & Random Horizontal Flip & 0.5                    \\
			\multirow{9}{*}{Trvial Augment}      & Shear X                & 0.0 -- 0.99             \\
			& Shear Y                & 0.0 -- 0.99             \\
			& Translate X            & 0 -- 32                 \\
			& Translate Y            & 0 -- 32                 \\
			& Rotate                 & $-135^\circ$ -- $+135^\circ$          \\
			& Brightness             & 0.01 -- 2.0             \\
			& \textbf{Contrast}      & \textbf{0.5 -- 1.8}     \\
			& Sharpness              & 0.5 -- 1.8              \\
			Advanced augmentation                & Random Erasing         & 0.1                    \\ \hline
		\end{tabular}
	\end{center}
\end{table}

Figure \ref{fig:TestCropSize} shows the EMR of the models at varying test crop size. We analyse models trained on a resolution of 224x224 and 512x512, again, trained on dacl1k, meta2+dacl1k and meta3+dacl1k.
Regarding the EMR on itself, it can be stated that the performance of the models trained with a train crop size of 224x224 is nearly identical to the performance of models trained with 512x512. 
Considering models trained with 512x512 images and tested on dacl1k, it can be stated that the best resolution for meta2+dacl1k is achieved at 560 while for meta3+dacl1k two peaks at a value of 528 and 624 with the same EMR are visible. The model trained on dacl1k shows its best performance at a test crop size of 528. 

\clearpage
\bibliographystyle{elsarticle-num} 
\bibliography{ref}

\end{document}